\pdfoutput=1
\documentclass[10pt,twocolumn,letterpaper]{article}

\usepackage[pagenumbers]{cvpr}

\usepackage{amsmath}
\usepackage{amssymb}
\usepackage{bm}
\usepackage{booktabs}
\usepackage{multirow}
\usepackage{xspace}
\usepackage{graphicx}
\usepackage{xcolor}
\usepackage{colortbl}

\definecolor{tabhead}{HTML}{E9EEF3}  % header rows: cool grey (cbg)
\definecolor{tabours}{HTML}{E1EEF7}  % LandWM rows: light blue (cblue)
\definecolor{tabbase}{HTML}{EAF4EE}  % baselines and references: light green (cgreen)
\definecolor{tabrule}{HTML}{6B7280}  % rules: neutral ink (cgray)
\arrayrulecolor{tabrule}

\newcommand{\method}{LandWM\xspace}

\newcommand{\zt}{\bm{z}_t}
\newcommand{\hphys}{\bm{h}}
\newcommand{\ufree}{\bm{u}}
\newcommand{\act}{\bm{a}}
\newcommand{\ctx}{\bm{c}}
\newcommand{\stat}{\bm{S}}
\newcommand{\obs}{\mathcal{O}}

\definecolor{cvprblue}{rgb}{0.21,0.49,0.74}
\usepackage[pagebackref,breaklinks,colorlinks,allcolors=cvprblue]{hyperref}

\title{Learning Land-Surface Hydrology World Model\\from Satellite Observations}

\author{%
Suya Li$^{1}$\qquad Zekun Qi$^{2}$\qquad Xiaochong Xu$^{1}$\qquad Duo Zhang$^{1}$\\[0.4em]
$^{1}$Beijing Forestry University\qquad $^{2}$Tsinghua University
}

\begin{document}
\twocolumn[{%
\renewcommand\twocolumn[1][]{#1}%
\maketitle
\begin{center}
  \captionsetup{type=figure}
  \includegraphics[width=\textwidth]{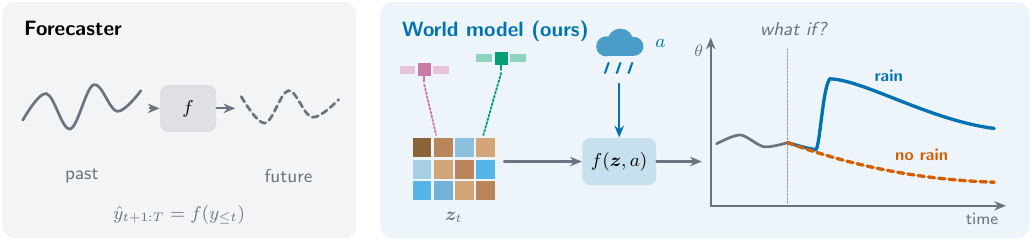}
  \captionof{figure}{\textbf{Forecasting versus world modelling.}
  (a) A forecaster regresses the future on the past.
  (b) \method learns a latent land state $\bm z_t$ from satellite observations and how it responds to an action $a$ (rain), so it can roll out what would happen under an alternative action.}
  \label{fig:teaser}
  \vspace{0.6em}
\end{center}%
}]
\begin{abstract}
A world model should tell us not only what will happen, but what would happen under a different action.
Learned Earth models now forecast remarkably well, yet forecast skill says little about what their state means or how it responds to a changed driver.
We build \method, a world model of land-surface water in which rain is the action, learned from seven years of SMAP and CYGNSS observations: physical observation operators tie its latent state to soil water, and a mass-conserving balance governs how that state responds to rain.
Over the U.S.\ Southern Great Plains, the model, never shown a soil-moisture value, tracks 28 in-situ probes as well as the operational SMAP Level-4 product, even from CYGNSS alone, and a storm added to the forcing of a station's cell wets its soil as natural storms wet the probes.
Replace either piece and the forecasts stay, but the meaning goes: with neural operators, only a probe fitted to the stations finds the soil water; with a generic recurrent transition, rain on a single cell barely moves it.
Yet when the same rain covers the whole region the two transitions swap roles, and the conserving one responds too weakly.
Grounding makes a world model readable; its response to actions must be tested under more than one kind of intervention.
\end{abstract}

\section{Introduction}
\label{sec:intro}

When rain falls on a field, where does the water go, and how long does it stay?
The answer decides whether a crop needs irrigation, whether the next storm runs off into a flood, and when a dry spell turns into drought~\cite{koster2004regions,mccoll2017global}.
It is also a question about alternatives: how wet would the soil be had the storm not come, and how fast will it dry if no rain falls for ten days?
Such questions are what \emph{world models} are for.
A world model learns how an environment responds to actions; it lets agents plan in imagination~\cite{ha2018world,hafner2019planet,hafner2025dreamerv3} and underlies foundation models of the physical world~\cite{nvidia2025cosmos} and of the Earth system~\cite{wang2026whatif}.

Machine learning has transformed how we forecast the Earth.
Learned weather models now rival operational systems~\cite{lam2023graphcast,price2025gencast,bodnar2025aurora}, some trained directly on observations~\cite{alexe2024graphdop,allen2025aardvark}; on land, recurrent networks forecast soil moisture and streamflow~\cite{fang2017prolongation,kratzert2019towards,nearing2024floods}, and emulators reproduce land-surface models~\cite{raoult2026ailand}.
Yet these models are forecasters: they learn to continue the past, and drivers such as rain enter, if at all, as one more input.
Whether it responds correctly when the driver \emph{changes} is rarely tested (\cref{fig:teaser}).

Forecast skill cannot settle this question.
A model can predict its observations well while holding a wrong picture of the process behind them~\cite{vafa2024evaluating,kang2024howfar,motamed2025physicsiq}, and a latent state learned from observations alone is not identifiable without further structure~\cite{locatello2019challenging,khemakhem2020ivae}.
The problem is sharpest in natural systems, whose state is never seen.
A satellite measures microwave brightness or the strength of a reflected radio signal, not the water in the soil~\cite{entekhabi2010smap,ruf2016cygnss}, and in-situ probes sample a vanishing fraction of the land.
Two questions follow: how can a model learn the response to an action without ever observing the state that responds, and how can we tell that it did?

Our answer to the first is to let physics fix what the state means.
We build \method, a world model of land-surface water in which precipitation is the action.
Its latent state holds water storages on a 9\,km grid, which change under a mass-conserving water balance, alongside free channels for whatever the balance leaves out.
The state is never compared with soil moisture.
Instead, each sensor is rendered from it by a differentiable physical forward model with a small bounded residual: an emission model~\cite{mo1982tauomega,mironov2009dielectric} for the L-band radiometer of SMAP and a coherent-reflection model~\cite{deroo1994bistatic} for the GNSS reflections of CYGNSS.
To explain what the satellites see, the storages must behave like soil water.
Cross-attention assimilates irregular observations from any sensor, and the model learns to predict future observations from rain alone over open-loop rollouts.
We train it on seven years of observations over the U.S.\ Southern Great Plains, 70.8 million SMAP and CYGNSS observation tokens, with IMERG precipitation~\cite{huffman2020imerg} as the action, MERRA-2 weather as context, and no reanalysis or retrieval as a target.

Our answer to the second is to let nature run the experiment.
Rain falls largely independently of the soil it lands on, so it acts as a natural intervention.
When we add a storm to the forcing of a station's cell, the wetting of its soil can be checked against what held-out probes record after natural storms of the same size; and when the satellite rain estimate misses a storm that a gauge records, the model's state can be checked against the probe.
In a twin experiment with known truth, we compare every model with the true counterfactual.

The results teach two lessons: forecast skill tells neither whether a world model can be read nor whether it responds to actions as the real system does; grounding decides the first, structure the second.
In the twin, every world model trained on observations learns the causal effect of rain better than a regression on rain history, and a probe fitted to the true state finds it in all of them; but only the physically grounded ones expose it in their state without labels.
Real data repeat this split.
The surface storage of \method, never shown a soil-moisture value, tracks 28 in-situ probes with a median correlation of 0.82, on par with SMAP Level-4 and above ERA5-Land and an LSTM trained on SMAP retrievals, also when only CYGNSS is assimilated and in a held-out test year, and a linear probe fitted to the stations reads it no better.
It is more than a retrieval or a memory of past rain: it improves on per-overpass retrievals of the same sensors, and where IMERG misses a storm that the station's gauge records, its state moves toward the probe, which no regression on the rain history can do.
Replace its physical observation operators by neural ones, and the model forecasts the satellites as well and its latent still holds the soil water (probe $R=0.81$), yet its state no longer tracks the probes ($R=0.32$): the knowledge is there, but only labels can find it.
Replace its conserving transition by a recurrent network, and the model forecasts nearly as well, yet a 20--50\,mm storm on a single cell raises its soil moisture by 0.002, against 0.11 at the probes and 0.13 for \method.
Spread the same storm over the whole region, however, and the roles swap: the recurrent model now wets its soil by 0.11, \method by only 0.035.
Each transition is right under one intervention and wrong under the other, a failure that neither forecast skill nor a single test reveals.

In summary, we formulate land-surface water as a world model with rain as the action, learned from satellite observations; we design \method, which grounds a mass-conserving latent state through physical observation operators; and we test the learned response, not forecast skill, through true counterfactuals in a twin and natural experiments on real data, showing that the observation operators decide whether a learned state is readable without labels, and the conserving transition whether it responds to local rain, and that a learned response must be tested under more than one kind of intervention.

\section{Related Work}
\label{sec:related}

\paragraph{World models and latent state-space models.}
Learned world models compress observations into a latent state and roll it forward under actions~\cite{ha2018world,hafner2019planet,hafner2025dreamerv3}.
In robotics, world models forecast the consequences of actions, in pixels or in a latent space, to guide vision--language--action policies~\cite{zhang2025dreamvla,sun2026vlajepa,zhang2026defi,zhang2026imagewam}.
Deep state-space models~\cite{krishnan2017structured} and latent ODEs~\cite{rubanova2019latentode} handle partial and irregular observations, and linear recurrent and structured state-space layers~\cite{gu2022s4,orvieto2023lru} provide long memory.
They mostly target agents with dense pixels; we address sparse observations, sensor-specific physics and a natural action.
The closest Earth-science use of the term, the Earth-system world model of Wang \etal~\cite{wang2026whatif}, learns from simulator trajectories with state edits as actions.
Earth-observation world models forecast vegetation indices under weather scenarios~\cite{luo2026eowm,iele2026vegsim} or image usability~\cite{albughdadi2026observability}; we ask instead whether the latent state can be read without labels.

\paragraph{Do world models learn the world?}
Good prediction does not imply a correct world model: generative sequence models can predict valid next moves while holding incoherent maps of the underlying world~\cite{vafa2024evaluating} or fitting orbits without Newtonian mechanics~\cite{vafa2025found}, and video generators often violate simple physical laws despite realistic frames~\cite{kang2024howfar,motamed2025physicsiq}, which motivates rewarding embodied video generation for physical realism~\cite{peng2026reworld}.
Vision--language models likewise still reason poorly about space~\cite{jia2026omnispatial}, and grounding them in 3D helps~\cite{qi2024shapellm,chen2026gsreasoner}.
Identifiability theory explains part of this: without inductive biases, latent factors cannot be recovered from observations~\cite{locatello2019challenging}, whereas suitable structure, such as conditioning variables, makes them identifiable~\cite{khemakhem2020ivae}.
Idealized perturbations test a learned weather model beyond skill~\cite{hakim2023dynamical}.
These studies have a known, simulated or idealized truth; ours is never observed, and physical observation operators and a natural action ground the latent.

\paragraph{Machine learning for weather and land.}
Global weather emulators trained on reanalysis~\cite{lam2023graphcast,price2025gencast,bodnar2025aurora} reach operational skill.
End-to-end systems that ingest observations include Aardvark~\cite{allen2025aardvark}, which is pre-trained on reanalysis, and GraphDOP~\cite{alexe2024graphdop,boucher2025coupled}, which is trained only on observations, develops Earth-system representations in its latent~\cite{lean2025graphdoprep} and inspires our query-able decoder; Earth-o1~\cite{gong2026eartho1} forecasts the atmosphere from observations without a grid.
On land, LSTMs predict SMAP soil moisture from meteorological forcing~\cite{fang2017prolongation,fang2020nrt}, catchment LSTMs predict streamflow~\cite{kratzert2019towards,nearing2024floods}, differentiable process models embed neural networks in hydrological structure~\cite{feng2022differentiable,shen2023differentiable}, mass-conserving LSTMs route water by construction~\cite{hoedt2021mclstm}, aiLand emulates a land-surface model before fine-tuning on observations~\cite{raoult2026ailand}, and process-guided learning retrieves soil moisture from SMAP brightness temperatures with a network pretrained on radiative-transfer simulations and fine-tuned to probes~\cite{feng2026pgml}.
These models use retrieved, modelled or in-situ soil moisture as targets, operate per pixel, or rely on a simulator; none learns spatial, action-conditioned dynamics from radiometric and reflectometry observations.

\paragraph{Learned data assimilation and observation operators.}
Score-based and latent-space assimilation~\cite{rozet2023sda,si2025latentensf,andry2025appa} learns priors from reanalysis and conditions them on observations.
Neural observation operators map land-model states to microwave brightness temperatures for assimilation~\cite{vural2026nnoperator}.
\method learns prior, assimilation and observation operators jointly from observations, and constrains the operators with physical forward models so that the latent keeps a physical meaning.
As in APHYNITY~\cite{yin2021aphynity} and physics-integrated VAEs~\cite{takeishi2021pivae}, a bounded neural correction leaves the physics to explain the data (\cref{sec:supp_sens}).
SMART~\cite{crow2011smart} and SM2RAIN~\cite{brocca2014sm2rain} correct rainfall from retrieved soil moisture, and differentiable hydrologic models update states or forcings~\cite{jamaat2025update}; the supplement tests this in a learned world model, as in weak-constraint 4D-Var~\cite{tremolet2006weak}.

\paragraph{GNSS reflectometry for land water.}
CYGNSS reflectivity tracks soil moisture~\cite{chew2018smap} and inundation~\cite{chew2018inundation,morris2019everglades} and is mapped to soil moisture pointwise~\cite{eroglu2019cygnss}; such static retrievals ignore dynamics, whereas we use specular points as observations of a dynamic state, as suits HydroGNSS~\cite{unwin2021hydrognss}.

\section{Method}
\label{sec:method}

\begin{figure*}[t]
  \centering
  \includegraphics[width=0.70\textwidth]{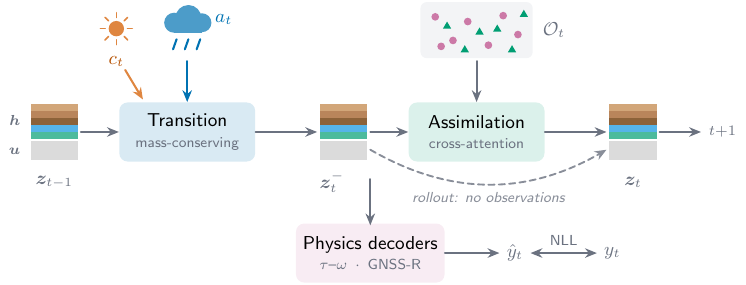}
  \caption{\textbf{One step of \method.}
  A mass-conserving transition advances the latent state, physical channels $\bm h$ and free memory $\bm u$, under rain $a_t$ and context $c_t$.
  Cross-attention assimilates the irregular observation tokens $\mathcal{O}_t$.
  Physics decoders score the observations under the prior $\bm z^-_t$.
  In open-loop rollouts, assimilation is skipped.}
  \label{fig:pipeline}
\end{figure*}

\subsection{Problem formulation}
We discretize a region into an $H\times W$ window of the 9\,km EASE-Grid~2.0~\cite{brodzik2012ease2} and time into steps of $\Delta t=6$\,h.
At step $t$ the model maintains a latent state $\zt\in\mathbb{R}^{H\times W\times d}$.
The action $\act_t\in\mathbb{R}^{H\times W}$ is the precipitation accumulated over $[t,t+\Delta t)$ from IMERG~\cite{huffman2020imerg}.
The context $\ctx_t$ collects the other meteorological forcing (radiation, temperature, humidity, wind, pressure) from MERRA-2~\cite{gelaro2017merra2}; it contains forcing fields only, never land-state variables.
$\stat$ denotes static attributes (terrain, soil texture, land cover, open-water occurrence).
The observations $\obs_t=\{o_k\}$ that arrive near step $t$ form an unordered set of tokens
$o_k=(m_k,\ \ell_k,\ \tau_k,\ \bm{g}_k,\ \bm{y}_k)$.
Each token holds the sensor identity $m_k$, the exact location $\ell_k$, the time offset $\tau_k\in[-3,3)$\,h from the state time, the acquisition geometry and quality indicators $\bm{g}_k$, and the measured value $\bm{y}_k$.
The state $\zt$ has received only the rain before $t$; observations go to the nearest state time, and \method is a state-space model
\begin{align}
  \text{transition:}\quad & \bm{z}^{-}_t = f_\theta(\bm{z}_{t-1}, \act_{t-1}, \ctx_{t-1}, \stat),\\
  \text{assimilation:}\quad & \zt = g_\phi(\bm{z}_t^{-}, \obs_t, \stat),\\
  \text{observation:}\quad & p_\psi(\bm{y}_k \mid \bm{z}^{-}_t, m_k, \ell_k, \bm{g}_k, \stat),
\end{align}
where $\bm{z}_t^{-}$ is the prior state produced by the transition (\cref{fig:pipeline}).
No term of the objective uses a reanalysis or retrieval product as a target.

\subsection{Latent state}
The latent is split as $\zt=[\hphys_t,\ufree_t]$.
The physical channels $\hphys_t$ are five non-negative water storages in millimetres: surface soil water $S_s$ in the top 5\,cm (so that $\theta_s=S_s/50\,\text{mm}$), root-zone soil water $S_r$, deep storage $S_g$, ponded surface water $S_w$ and vegetation water $V$.
The 96 free channels $\ufree_t$ represent unmodelled factors such as roughness, vegetation structure and management.
The observation operators (\cref{sec:decoders}) fix the meaning and scale of $\hphys$, up to their bounded learned parameters, which lets us compare $\theta_s$ directly with in-situ probes never used for training.

\subsection{Transition}
\label{sec:transition}
\paragraph{Water balance by partition.}
A convolutional network with a $3\times3$ receptive field reads $(\bm{z}_{t-1},\act_{t-1},\ctx_{t-1},\stat)$ and outputs flux logits for each cell.
The fluxes are applied in a fixed order:
(1) precipitation and ponded water infiltrate up to the remaining capacity of the surface layer;
(2) the excess is split into ponded water, runoff and evaporation;
(3) the surface layer into remaining water, evapotranspiration and percolation;
(4) the root zone into remaining water, transpiration and drainage, with any overflow becoming runoff;
(5) the deep storage receives drainage and loses baseflow.
Each split is a softmax over fractions of the current storage, so storages stay non-negative, outflows are limited by the water available, and water is conserved up to the explicit losses (evapotranspiration, runoff, baseflow).
Layer capacities (effective porosity times depth) are predicted from $\stat$.
Soils hold water that neither plants nor gravity remove~\cite{vangenuchten1980closed}, so steps 3--4 act only on the water above a residual content $\theta_{\mathrm{res}}(\stat)\in[0.02,0.15]$; below it, a dry layer stops losing water.
Vegetation water follows bounded leaky dynamics and is not part of the balance.
The physical storages exchange no water between cells; spatial context enters only through the flux network and the free channels.

\paragraph{Multi-timescale memory.}
Free channels evolve as
\begin{equation}
\ufree_t=\mathrm{Mix}\big(\bm{\lambda}(\stat)\odot\ufree_{t-1}+(1-\bm{\lambda}(\stat))\odot\Delta\ufree_t\big).
\end{equation}
The decay rates $\bm{\lambda}$ are learned per channel and modulated by $\stat$~\cite{orvieto2023lru}; their time scales are initialized log-uniformly between 6\,h and one year.
$\mathrm{Mix}$ is a depthwise $5\times5$ convolution followed by a pointwise convolution, which exchanges information between neighbouring cells.
This memory keeps rollouts of hundreds of steps stable.

\subsection{Assimilation by cross-attention}
Each observation token is embedded by a sensor-specific MLP.
The MLP reads $\bm{y}_k$, $\bm{g}_k$ and Fourier features of the sub-cell offset and of $\tau_k$, and a learned sensor embedding is added.
Each grid cell forms a query from its prior state and static attributes and attends, with four heads, to the tokens that fall inside the cell.
Two $3\times3$ convolutions spread the aggregated message, together with the log token count, to neighbouring cells.
A gated network then produces the increment: storages are updated multiplicatively, $S\leftarrow S\exp\!\big(\tfrac12\tanh\Delta S\big)$, so they stay positive and change by a bounded factor per step; free channels are updated additively.
The gate starts nearly closed, so training starts close to an open-loop model.

\subsection{Physics-plus-residual observation operators}
\label{sec:decoders}
Each sensor has a decoder that interpolates the latent bilinearly to the observation location.
It outputs the location and scale of a Student-$t$ likelihood ($\nu=5$), which is robust to radio-frequency interference and outliers.
Decoders see only geometry that is fixed before the measurement.
CYGNSS products also report quantities computed from the measured signal itself, such as signal-to-noise and coherence ratios, which together with the geometry explain 96\% of the variance of the reflectivity.
They enter the assimilation tokens, but never a decoder, which would otherwise predict the observation from itself.

\paragraph{SMAP.}
The physical part is the zeroth-order $\tau$--$\omega$ model~\cite{mo1982tauomega}.
The soil permittivity comes from $\theta_s$ through the Mironov dielectric model~\cite{mironov2009dielectric}, followed by Fresnel reflection with a roughness correction.
Vegetation optical depth grows linearly with $V$ plus a learned term.
Single-scattering albedo and roughness are learned within physical bounds.
The effective temperature is the 2\,m air temperature at the state time plus a learned correction of at most $\pm5$\,K.

\paragraph{CYGNSS.}
The physical part is coherent reflection~\cite{deroo1994bistatic}:
\begin{equation}
\Gamma=|R_{lr}(\theta_i,\varepsilon(\theta_s))|^2\,e^{-(2k\sigma\cos\theta_i)^2}\,e^{-2\tau_v/\cos\theta_i}.
\end{equation}
It is mixed with open-water reflection according to the inundated fraction $f_w=1-e^{-S_w/d}$, where the depth scale $d$ is learned.
Roughness $\sigma$ and attenuation $\tau_v$ come from $V$ and the free channels.
An embedding per receiver--transmitter pair absorbs instrument biases.
For both sensors, a neural residual bounded to $\pm2$\,K and $\pm1$\,dB absorbs calibration and unmodelled effects, so the physics, not the residual, must explain the large signals (removing it changes the real-data results little; \cref{sec:supp_sens}).

\subsection{Training objective}
For a window starting at $t_0$, the model filters for $T_f$ steps and then rolls out open loop for $H$ steps using actions and context only.
Every observation is scored under the state \emph{before} it is assimilated.
The objective is thus always a prediction, and the model cannot lower it by copying its input:
\begin{align}
\mathcal{L} = \sum_{m}\frac{1}{N_m}\Big[&\sum_{t=t_0+T_b}^{t_0+T_f}\ \sum_{o_k\in\obs^m_t}-\log p_\psi(\bm{y}_k\mid\bm{z}^{-}_t)\nonumber\\
&+\sum_{h=1}^{H}\ \sum_{o_k\in\obs^m_{t_0+T_f+h}}-\log p_\psi\big(\bm{y}_k\mid\hat{\bm{z}}_{t_0+T_f+h}\big)\Big],
\end{align}
where $\hat{\bm{z}}$ denotes open-loop states, $T_b$ a burn-in period, and $N_m$ the number of observations of sensor $m$ in each phase, so that each sensor and phase contributes equally.
The rollout horizon follows a curriculum from one day to ten days.
In training, each sensor is withheld from assimilation with probability 0.2 (still scored), for a sensor-agnostic latent.

\section{Experiments}
\label{sec:experiments}

\subsection{Twin experiment with known truth}
\label{sec:twin}
\paragraph{Synthetic world.}
Real soil moisture is never observed on the model grid, so we first test the method in a synthetic world.
The truth model deliberately differs from the transition of \method in functional form (two soil layers, power-law infiltration, nonlinear drainage, threshold ponding), and its parameters are random fields the model never sees; precipitation and context mimic the Southern Great Plains (details in the supplement, \cref{sec:supp_twin}).
Observations are generated with the forward models of \cref{sec:method}, the decoders' own family (\cref{sec:supp_ms} uses operators outside it): SMAP-like brightness temperatures on swaths with a 2--3-day revisit ($\sigma=1.5$\,K) and CYGNSS-like reflectivities at random specular points with per-receiver/transmitter biases ($\sigma=1$\,dB).
We use a $64\times64$ domain with 6\,h steps and three years of data, holding out the last 25\% in time.

\paragraph{Readability and information.}
We compare the model's surface storage with the true surface soil moisture by per-cell temporal correlation, averaged in space, both while assimilating and in 10-day open-loop rollouts driven only by precipitation and context.
We ask whether the designated surface-storage channel equals soil moisture with no fitting (\emph{readability}), and whether a linear (ridge) probe fit on the training period decodes it from the latent of any model, black boxes and the Transformer included (\emph{information}).
As a model-free reference, the same probe reads antecedent-precipitation features (causal filters of rain and potential evaporation, 6\,h to 30\,d, and static fields).

\paragraph{True counterfactuals.}
The twin also gives us what no observational benchmark can: the \emph{true counterfactual}.
After the last assimilation step, we remove all precipitation and re-simulate the truth from its true state.
We then compare the model's zero-rain rollout with the true counterfactual: the causal effect of the observed rain, $\theta(\text{actual})-\theta(\text{no rain})$, by correlation and regression slope (1 means the right magnitude), and the per-cell drydown time scale $\tau$.

\paragraph{Models.}
The ablations remove the residual water content, replace the physical decoders by neural ones or the conserving transition by a GRU, withhold the action, drop the rollout loss, or combine neural decoders and a GRU (black box); the autoregressive observation-token Transformer (History Transformer) is a further baseline.
Seeds, each drawing its own synthetic world, are given in \cref{tab:twin}; the other conserving variants keep the residual content.

\begin{table*}[t]
\centering
\caption{\textbf{Twin experiment.}
State $R$: correlation of surface soil moisture with the truth in 10-day open-loop rollouts, read from the designated storage with no fitting (direct) or by a ridge probe fit on the training period.
Rollout NLL: per-token NLL of SMAP observations in the same rollouts.
Rain effect: all rain is removed after the last assimilation step; correlation and regression slope of the model's $\theta(\text{actual})-\theta(\text{no rain})$ on the true effect, and rank correlation $\rho$ of the per-cell drydown $\tau$ with the truth.
Mean $\pm$ std over seeds, each with its own synthetic world (three for \method and no residual content, two for neural decoder, GRU and black box).
Full tables in the supplement (\cref{tab:twin_full,tab:twin_cf}).}
\label{tab:twin}
\resizebox{\textwidth}{!}{% generated by code/scripts/make_twin_table.py -- do not edit by hand
\begin{tabular}{l cc c ccc ccc}
\toprule
\rowcolor{tabhead}  & \multicolumn{2}{c}{State $R$ (rollout)} & Rollout & \multicolumn{3}{c}{Rain effect, direct} & \multicolumn{3}{c}{Rain effect, probe} \\
\cmidrule(lr){2-3}\cmidrule(lr){5-7}\cmidrule(lr){8-10}
\rowcolor{tabhead} Model & direct & probe & SMAP NLL & $R$ & slope & $\tau$ $\rho$ & $R$ & slope & $\tau$ $\rho$ \\
\midrule
\rowcolor{tabours} \method & 0.83$\pm$0.03 & 0.84$\pm$0.03 & 3.90$\pm$0.02 & 0.99$\pm$0.00 & 1.08$\pm$0.03 & 0.97$\pm$0.01 & 0.99$\pm$0.00 & 0.97$\pm$0.01 & 0.98$\pm$0.00 \\
\quad no residual content & 0.80$\pm$0.04 & 0.83$\pm$0.03 & 3.97$\pm$0.05 & 0.98$\pm$0.01 & 1.15$\pm$0.05 & 0.75$\pm$0.07 & 0.98$\pm$0.01 & 0.97$\pm$0.01 & 0.78$\pm$0.09 \\
\quad neural decoder & 0.18$\pm$0.01 & 0.81$\pm$0.04 & 3.95$\pm$0.07 & 0.04$\pm$0.08 & 0.02$\pm$0.03 & 0.49$\pm$0.35 & 0.97$\pm$0.01 & 0.95$\pm$0.02 & 0.90$\pm$0.03 \\
\quad GRU transition & 0.84$\pm$0.03 & 0.82$\pm$0.03 & 3.82$\pm$0.00 & 0.99$\pm$0.00 & 1.31$\pm$0.11 & 0.99$\pm$0.00 & 1.00$\pm$0.00 & 0.99$\pm$0.00 & 0.98$\pm$0.00 \\
\quad no action & 0.13 & 0.10 & 6.44 & -0.00 & -0.00 & -0.10 & -0.00 & -0.00 & -0.11 \\
\quad no rollout loss & 0.75 & 0.72 & 7.19 & 0.93 & 1.34 & 0.91 & 0.94 & 0.97 & 0.62 \\
\quad black box & -0.10$\pm$0.15 & 0.81$\pm$0.05 & 3.86$\pm$0.00 & -0.19$\pm$0.44 & -0.03$\pm$0.09 & 0.33$\pm$0.08 & 0.99$\pm$0.00 & 1.01$\pm$0.01 & 0.97$\pm$0.00 \\
\rowcolor{tabbase} History Transformer & -- & 0.72 & 4.16 & -- & -- & -- & 0.89 & 0.80 & 0.95 \\
\midrule
\rowcolor{tabbase} Rain-history ridge (no model) & -- & 0.75$\pm$0.03 & -- & -- & -- & -- & 0.79$\pm$0.04 & 0.65$\pm$0.07 & 0.40$\pm$0.02 \\
\bottomrule
\end{tabular}
}
\end{table*}

\paragraph{Forecasts do not reveal the state.}
Forecast skill does not separate the models (\cref{tab:twin}): all models that receive the action and are trained on rollouts predict observations about equally well (rollout SMAP NLL 3.82--4.16, against 6.44 without the action).
Only with physical observation operators does the surface channel track soil moisture without fitting (rollout $R$ 0.83 for \method and 0.84 with a GRU transition, against 0.18 with a neural decoder and $-0.10$ for the black box).
A probe recovers soil moisture from every model that receives the action (rollout $R$ 0.72--0.84), so the information is there but, without grounding, not readable.

\paragraph{The response to rain is learned, but readable only with grounding.}
Read through a probe, every model that receives the action recovers the causal effect of rain (effect $R\geq0.89$, slope 0.80--1.01), better than the rain-history ridge (slope $0.65\pm0.07$, drydown rank correlation $0.40$): learning from observations captures the response to the action, not only its correlation with the past.
Read directly, only the grounded models expose it: effect $R$ is 0.99 for \method and with a GRU transition, against 0.04 with a neural decoder and $-0.19$ for the black box.

\paragraph{Residual content and transition shape the drydown.}
Without the residual content the soil keeps drying below the content at which the true losses stop, which the exponential fit reads as a slow decay (\cref{tab:twin_cf}; median $\tau$ 4.8 against 2.7\,days, rank correlation $0.75\pm0.07$).
With it, \method matches the drydown (3.0\,days, rank correlation $0.97\pm0.01$) and the size of the effect (slope $1.08\pm0.03$ against $1.15\pm0.05$).
The GRU transition with the same decoders also matches the drydown (2.6\,days, 0.99) but overestimates the effect more (slope $1.31\pm0.11$), so in the twin grounding comes from the observation operators.

\paragraph{The split survives a wrong truth and wrong operators.}
This twin shares two assumptions with \method: water moves only vertically, and the observations come from the decoders' own forward models.
We break each in turn, keeping the worlds, forcing and noise (\cref{tab:twin_robust}).
In a second twin, the truth adds run-on to the lowest neighbour, lateral subsurface flow, and irrigation of 20\% of the cells that is not part of the action (\cref{sec:supp_mm}).
In a third, the observations are generated with operators outside the decoders' family: another dielectric model, polarization mixing, moisture-dependent roughness, a lagged vegetation opacity and an incoherent CYGNSS component, which change the brightness temperatures by 6.4 and 13.6\,K (rms, H and V) and the reflectivity by 2.1\,dB, well above the noise (\cref{sec:supp_ms}).
The observation likelihood still barely separates the models, and a probe still reads soil moisture (0.76--0.79; History Transformer 0.69 and 0.71) and the effect of rain ($R\geq0.96$) from all of them.
Read directly, only the physical decoders expose both: state $R$ 0.78 and 0.81 for \method, against 0.20 and 0.00 with a neural decoder and $-0.26$ and $-0.18$ for the black box.
Wrong operators cost magnitudes rather than readability (effect slope $1.25\pm0.02$ against $1.08\pm0.03$): grounding needs the right family, not exact operators.

\begin{table*}[t]
\centering
\caption{\textbf{Twins that break the model's assumptions.}
Left: the truth adds lateral run-on, lateral subsurface flow and unobserved irrigation. Right: observations are generated by operators outside the decoders' family. Columns as in \cref{tab:twin}; full tables in \cref{tab:twin_mm,tab:twin_ms}.}
\label{tab:twin_robust}
\small
% generated by arxiv/scripts/make_arxiv_tables.py from tables/{mm,ms}/twin_main.tex -- do not edit by hand
\begin{tabular}{l cccc cccc}
\toprule
\rowcolor{tabhead} & \multicolumn{4}{c}{Lateral flow and irrigation} & \multicolumn{4}{c}{Mis-specified operators} \\
\cmidrule(lr){2-5}\cmidrule(lr){6-9}
\rowcolor{tabhead} & \multicolumn{2}{c}{State $R$} & \multicolumn{2}{c}{Effect, direct} & \multicolumn{2}{c}{State $R$} & \multicolumn{2}{c}{Effect, direct} \\
\rowcolor{tabhead} Model & direct & probe & $R$ & slope & direct & probe & $R$ & slope \\
\midrule
\rowcolor{tabours} \method & 0.78$\pm$0.04 & 0.76$\pm$0.03 & 0.97$\pm$0.00 & 0.98$\pm$0.06 & 0.81$\pm$0.03 & 0.81$\pm$0.03 & 0.98$\pm$0.00 & 1.25$\pm$0.02 \\
\quad neural decoder & 0.20 & 0.77 & 0.45 & 0.29 & 0.00 & 0.77 & 0.05 & 0.03 \\
\quad GRU transition & 0.80 & 0.77 & 0.97 & 1.21 & 0.79 & 0.79 & 0.99 & 1.33 \\
\quad black box & -0.26 & 0.76 & -0.23 & -0.13 & -0.18 & 0.79 & -0.03 & -0.01 \\
\rowcolor{tabbase} History Transformer & -- & 0.69 & -- & -- & -- & 0.71 & -- & -- \\
\bottomrule
\end{tabular}

\end{table*}

\subsection{Real-data setup}
\label{sec:real}
\paragraph{Region and period.}
We study the U.S.\ Southern Great Plains (104$^\circ$W--94$^\circ$W, 30$^\circ$N--38$^\circ$N; a $95\times108$ window of the 9\,km grid), which lies inside CYGNSS coverage, hosts dense in-situ networks and spans a climate gradient from semi-arid west to humid east.
We train on August 2018--December 2023, validate on 2024, and test on January--September 2025, the current extent of IMERG Final.

\paragraph{Observations.}
SMAP enhanced L1C brightness temperatures (H and V polarization)~\cite{entekhabi2010smap} and CYGNSS Level-1 specular points~\cite{ruf2016cygnss} inside the region.
After quality control, we have 46.9\,M SMAP and 23.9\,M CYGNSS tokens, observing 45\% and 53\% of the 6\,h steps.

\paragraph{Independent evaluation data.}
Soil moisture and precipitation from the U.S.\ Climate Reference Network~\cite{diamond2013uscrn} (11 stations) and the Soil Climate Analysis Network~\cite{schaefer2007scan} (17 stations).
We score the surface (5\,cm) probes; their gauges also time natural storms and check IMERG.
None of these data are used for training or model selection.

\paragraph{Baselines.}
We compare with oracle persistence of the station value, per-overpass retrievals (the radiative-transfer SMAP Level-3, the process-guided PGML~\cite{feng2026pgml} and CYGNSS Level-3), a per-pixel LSTM trained on SMAP retrievals~\cite{fang2017prolongation}, the History Transformer, and two hand-built grounded dynamical models: the land-surface models of SMAP Level-4~\cite{reichle2017smapl4}, which assimilates SMAP brightness temperatures, and ERA5-Land~\cite{munozsabater2021era5land}, forced by reanalysis; and the rain-history ridge of the twin, fitted to each probe on the training period.
Differentiable process models and land-model emulators~\cite{feng2022differentiable,raoult2026ailand} are trained on soil-moisture products or land-model output, which our setting excludes.

\paragraph{Metrics.}
Soil moisture is read directly from the state, $\theta_s=S_s/50\,$mm in the station's 9\,km cell, without any calibration to in-situ data.
Because a 9\,km cell and a 5\,cm probe differ in level, we report per-station correlation, unbiased RMSE and the correlation of anomalies from a 31-day moving mean, summarised by the median over stations.

\paragraph{Tasks.}
\emph{Filtering}: $\theta_s$ while observations are assimilated, scored before each step's observations enter.
\emph{Action-conditioned forecasting}: from a filtered state and observed precipitation, predict SMAP and CYGNSS observations and station soil moisture 1--10 days ahead.
\emph{Action grounding}: degradation when actions are zeroed, shuffled in time, or swapped between years.
\emph{Response to rain}: we add synthetic rain pulses to the forcing of the station's cell in a filtered state at dry times (over the whole domain in \cref{sec:local}) and compare the increment of soil moisture with that of isolated natural storms of the same size at the probes, grouped by amount and antecedent wetness.
SMAP Level-4 assimilates SMAP brightness temperatures and gauge-corrected precipitation, so it enters the filtering comparison only.

\subsection{Results}

\begin{figure*}[t]
  \centering
  \includegraphics[width=\textwidth]{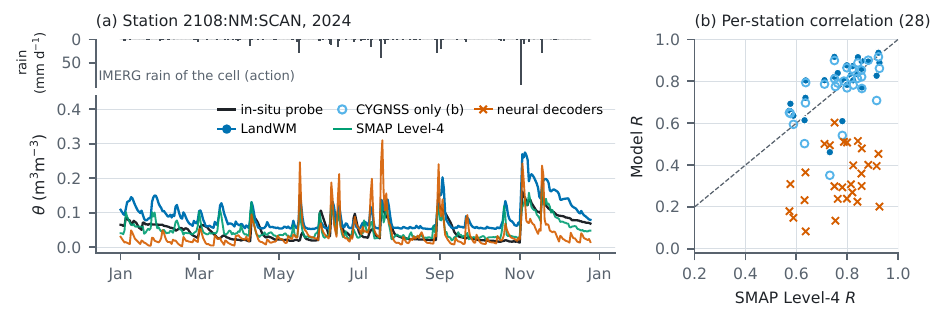}
  \caption{\textbf{Station soil moisture on the validation year} (seed 0).
  (a) Daily means at the station with the median correlation of \method among stations with a complete record; top: the IMERG rain of the cell, the action.
  (b) Per-station correlation against that of SMAP Level-4, on the same samples: \method assimilating both sensors or CYGNSS only, and the same model with neural decoders, which forecasts the observations equally well.}
  \label{fig:real}
\end{figure*}

\begin{table*}[t]
\centering
\caption{\textbf{Station soil moisture} on the validation year (median over 28 USCRN and SCAN stations).
Baselines are scored on the samples at which \method is scored; on the LSTM's daily (12\,UTC) samples \method obtains $R=0.82$.
Forecasts start from a filtered state with observed rain; oracle persistence keeps the station's value at the forecast start.
Response to rain: median increment of surface soil moisture after synthetic pulses of 10--20 and 20--50\,mm added to a filtered state; for the probes, after isolated natural storms of the same size (gauge rain).
The rain-history ridge is fitted to each station on the training period; \method never sees the stations.
Mean $\pm$ std over three seeds (\method, GRU transition) or two (other ablations).}
\label{tab:real_sm}
\resizebox{\textwidth}{!}{% generated by code/scripts/make_real_table.py -- do not edit by hand
\begin{tabular}{l ccc ccc cc}
\toprule
\rowcolor{tabhead}  & \multicolumn{3}{c}{Filtering} & \multicolumn{3}{c}{Forecast $R$, lead (days)} & \multicolumn{2}{c}{Response to rain} \\
\cmidrule(lr){2-4}\cmidrule(lr){5-7}\cmidrule(lr){8-9}
\rowcolor{tabhead} Model & $R$ & ubRMSE & anomaly $R$ & 1 & 5 & 10 & 10--20\,mm & 20--50\,mm \\
\midrule
\rowcolor{tabours} \method & 0.82$\pm$0.00 & 0.045$\pm$0.001 & 0.71$\pm$0.01 & 0.82$\pm$0.01 & 0.82$\pm$0.01 & 0.80$\pm$0.01 & 0.069$\pm$0.005 & 0.128$\pm$0.010 \\
\quad GRU transition & 0.81$\pm$0.01 & 0.053$\pm$0.006 & 0.71$\pm$0.01 & 0.81$\pm$0.00 & 0.78$\pm$0.01 & 0.79$\pm$0.01 & 0.001$\pm$0.000 & 0.002$\pm$0.000 \\
\quad neural decoder & 0.32$\pm$0.02 & 0.069$\pm$0.000 & 0.48$\pm$0.00 & 0.25$\pm$0.05 & 0.35$\pm$0.01 & 0.31$\pm$0.08 & -- & -- \\
\quad no action & 0.76$\pm$0.00 & 0.047$\pm$0.001 & 0.57$\pm$0.05 & 0.79$\pm$0.01 & 0.70$\pm$0.00 & 0.63$\pm$0.00 & -- & -- \\
\quad no rollout loss & 0.82$\pm$0.00 & 0.049$\pm$0.001 & 0.71$\pm$0.01 & 0.82$\pm$0.01 & 0.77$\pm$0.03 & 0.73$\pm$0.05 & 0.087$\pm$0.004 & 0.151$\pm$0.007 \\
\midrule
\rowcolor{tabbase} Persistence (oracle) & -- & -- & -- & 0.98 & 0.70 & 0.47 & -- & -- \\
\rowcolor{tabbase} Probes, natural storms & -- & -- & -- & -- & -- & -- & 0.062 & 0.108 \\
\rowcolor{tabbase} Rain-history ridge (station-fit) & 0.79 & 0.044 & 0.71 & 0.78 & 0.78 & 0.78 & -- & -- \\
\rowcolor{tabbase} LSTM on SMAP L3 & 0.72 & 0.050 & 0.66 & -- & -- & -- & -- & -- \\
\rowcolor{tabbase} SMAP Level-4 & 0.79 & 0.045 & 0.68 & -- & -- & -- & -- & -- \\
\rowcolor{tabbase} ERA5-Land & 0.71 & 0.052 & 0.61 & -- & -- & -- & -- & -- \\
\bottomrule
\end{tabular}
}
\end{table*}

\begin{table*}[t]
\centering
\caption{\textbf{Observation forecasts and action grounding} on the validation year.
Rollout NLL: per-token NLL of SMAP and CYGNSS observations over 10-day open-loop rollouts with the observed rain.
$\Delta$NLL: increase when the rain is set to zero, shuffled in time, or replaced by the same season of an earlier year (mean of one and two years).}
\label{tab:real_obs}
\small
\resizebox{\textwidth}{!}{% generated by code/scripts/make_real_table.py -- do not edit by hand
\begin{tabular}{l cc ccc ccc}
\toprule
\rowcolor{tabhead}  & \multicolumn{2}{c}{Rollout NLL} & \multicolumn{3}{c}{SMAP $\Delta$NLL} & \multicolumn{3}{c}{CYGNSS $\Delta$NLL} \\
\cmidrule(lr){2-3}\cmidrule(lr){4-6}\cmidrule(lr){7-9}
\rowcolor{tabhead} Model & SMAP & CYGNSS & zero & shuffle & swap & zero & shuffle & swap \\
\midrule
\rowcolor{tabours} \method & 5.41$\pm$0.01 & 2.05$\pm$0.01 & 3.31$\pm$0.10 & 1.40$\pm$0.01 & 2.65$\pm$0.03 & 0.29$\pm$0.01 & 0.14$\pm$0.00 & 0.26$\pm$0.00 \\
\quad GRU transition & 5.37$\pm$0.02 & 2.08$\pm$0.01 & 2.98$\pm$0.06 & 1.42$\pm$0.01 & 2.50$\pm$0.02 & 0.23$\pm$0.01 & 0.13$\pm$0.00 & 0.22$\pm$0.00 \\
\quad neural decoder & 5.40$\pm$0.02 & 2.06$\pm$0.00 & 3.49$\pm$0.00 & 1.47$\pm$0.04 & 2.79$\pm$0.00 & 0.28$\pm$0.01 & 0.14$\pm$0.00 & 0.25$\pm$0.01 \\
\quad no action & 6.02$\pm$0.05 & 2.11$\pm$0.01 & -- & -- & -- & -- & -- & -- \\
\quad no rollout loss & 7.40$\pm$0.76 & 2.44$\pm$0.12 & 3.28$\pm$0.18 & 0.97$\pm$0.27 & 2.06$\pm$0.33 & 0.32$\pm$0.16 & 0.11$\pm$0.04 & 0.21$\pm$0.08 \\
\quad black box & 5.34 & 2.07 & 3.06 & 1.47 & 2.53 & 0.24 & 0.13 & 0.22 \\
\rowcolor{tabbase} History Transformer & 6.18 & 2.16 & 3.77 & 1.91 & 3.45 & 0.26 & 0.14 & 0.25 \\
\rowcolor{tabbase} \quad best validation checkpoint & 5.73 & 2.26 & 3.64 & 1.60 & 3.05 & 0.28 & 0.12 & 0.23 \\
\bottomrule
\end{tabular}
}
\end{table*}

\begin{table}[t]
\centering
\caption{\textbf{Held-out sensor} on the validation year: \method (three seeds) with one sensor never assimilated. Columns as in \cref{tab:real_sm}.}
\label{tab:heldout}
\small
\resizebox{\linewidth}{!}{% generated by code/scripts/make_heldout_table.py -- do not edit by hand
\begin{tabular}{l c ccc c}
\toprule
\rowcolor{tabhead} Assimilated & seeds & $R$ & ubRMSE & anomaly $R$ & 10-day forecast $R$ \\
\midrule
\rowcolor{tabours} SMAP and CYGNSS & 3 & 0.82$\pm$0.00 & 0.045$\pm$0.001 & 0.71$\pm$0.01 & 0.80$\pm$0.01 \\
SMAP only & 3 & 0.82$\pm$0.00 & 0.046$\pm$0.001 & 0.71$\pm$0.01 & 0.80$\pm$0.00 \\
CYGNSS only & 3 & 0.80$\pm$0.00 & 0.046$\pm$0.002 & 0.67$\pm$0.01 & 0.78$\pm$0.01 \\
\bottomrule
\end{tabular}
}
\end{table}

\paragraph{Soil moisture.}
The model has never seen a soil-moisture value, yet its storage $S_s$ tracks the probes (\cref{tab:real_sm,fig:real}): on the same samples, the median correlation is 0.82 in each seed, against 0.79 for SMAP Level-4, 0.71 for ERA5-Land and 0.72 for the LSTM on SMAP retrievals (anomaly correlation 0.71 against 0.68, 0.61 and 0.66).
Paired over stations (Wilcoxon, per seed, uncorrected), it is better than ERA5-Land and the LSTM ($p\le0.001$) and than SMAP Level-4 in anomaly correlation (20--22 of 28 stations, $p\le0.01$).
With observed precipitation, 10-day forecasts keep $R=0.80$, where oracle persistence of the probe value has fallen to 0.47.
Even level and amplitude, which correlation ignores, are close (median bias $-0.02$ to 0.00, std ratio 0.98--1.16; \cref{sec:supp_retr}).
CYGNSS alone suffices (\cref{tab:heldout}): its specular points give $R=0.80$ and anomaly $R=0.67$, indistinguishable from SMAP Level-4, which assimilates SMAP (better at 14--16 of 28 stations, $p\ge0.34$), and better than ERA5-Land and the LSTM.
Dropping CYGNSS instead changes nothing, because SMAP observes every station cell at least every two days and leaves it no gaps to fill here (\cref{sec:supp_extra}).

\begin{table*}[t]
\centering
\caption{\textbf{Soil-moisture change on days when IMERG and the station's gauge disagree} (validation and test years). $\Delta\theta$: median change from before the day to the mean of the next day; wet: fraction of events with $\Delta\theta>0.02$. \method: mean over three seeds; closer: events at which its change is closer to the probe's than the ridge's (range over seeds), and two-sided Wilcoxon $p$ (uncorrected).}
\label{tab:mismatch}
\resizebox{\textwidth}{!}{% generated by code/scripts/make_mismatch_table.py -- do not edit by hand
\begin{tabular}{r l r cc cc cc cc}
\toprule
\rowcolor{tabhead}  & & & \multicolumn{2}{c}{Probe} & \multicolumn{2}{c}{Rain-history ridge} & \multicolumn{2}{c}{\method} & \multicolumn{2}{c}{\method vs.\ ridge} \\
\cmidrule(lr){4-5}\cmidrule(lr){6-7}\cmidrule(lr){8-9}\cmidrule(lr){10-11}
\rowcolor{tabhead} Threshold & Days & $n$ & $\Delta\theta$ & wet & $\Delta\theta$ & wet & $\Delta\theta$ & wet & closer & $p$ \\
\midrule
10\,mm & IMERG missed & 36 & +0.052 & 0.78 & -0.001 & 0.11 & +0.009 & 0.39 & 26--28/36 & 0.001--0.002 \\
10\,mm & IMERG false alarm & 54 & -0.003 & 0.22 & +0.037 & 0.70 & +0.030 & 0.60 & 31--36/54 & 0.024--0.198 \\
10\,mm & both $\ge$ threshold & 462 & +0.066 & 0.82 & +0.047 & 0.85 & +0.046 & 0.76 & 208--215/454 & 0.304--0.403 \\
\midrule
20\,mm & IMERG missed & 11 & +0.085 & 0.91 & -0.002 & 0.18 & +0.008 & 0.45 & 9--10/11 & 0.003--0.007 \\
20\,mm & IMERG false alarm & 19 & +0.000 & 0.11 & +0.068 & 0.95 & +0.038 & 0.70 & 14--15/19 & 0.002--0.009 \\
\bottomrule
\end{tabular}
}
\end{table*}

\paragraph{More than a retrieval or a memory of rain.}
Reading $\theta_s$ is a retrieval regularized by dynamics; what do the dynamics add?
On the samples of per-overpass retrievals of the same sensors, \method beats SMAP Level-3 ($R=0.77$ against 0.72, 17 of 20 stations) and, from CYGNSS alone, CYGNSS Level-3 (0.82 against 0.61, 25--26 of 26), and it matches PGML, a radiative-transfer-guided network fine-tuned to in-situ probes of these networks in earlier years (0.82 against 0.81; \cref{tab:retrieval}).
A regression on rain history, fitted to the probes, also comes close to \method at the stations ($R=0.79$), so we look where the two must part: days when IMERG and the station's gauge disagree (\cref{tab:mismatch}, \cref{sec:supp_mismatch}).
After 36 storms of at least 10\,mm that IMERG missed, the probes wet and the ridge cannot, but the state of \method, driven by the same IMERG rain yet also seeing the satellites, comes closer to the probe at 26--28 of them ($p\le0.002$); after 54 storms that IMERG invented, it does so at 31--36.
The separation grows with the size of the disagreement (9--10 of 11 missed storms of at least 20\,mm), and on days when IMERG and the gauge agree the two are equally close to the probe: the state departs from the rain history exactly where the satellites contradict it.

\paragraph{Observations and actions.}
Physical decoders cost little likelihood: rollout NLL is 5.41 for SMAP and 2.05 for CYGNSS (std 0.01), against 5.34 and 2.07 for the black box and 6.18 and 2.16 for the History Transformer (\cref{tab:real_obs}).
Both depend on the rain: zeroing it raises the SMAP NLL by $3.31\pm0.10$ (black box 3.06), the rain of another year by $2.65\pm0.03$ (2.53) and shuffling it by $1.40\pm0.01$ (1.47): the real rain beats a plausible one, and any rain beats none.

\begin{figure*}[t]
  \centering
  \includegraphics[width=\textwidth]{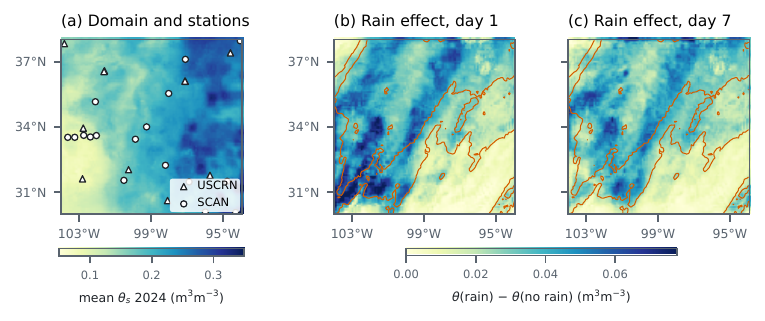}
  \caption{\textbf{\method on real data} (seed 0, validation year).
  (a) Mean filtered surface soil moisture $\theta_s=S_s/50$\,mm and the 28 in-situ stations.
  (b, c) Counterfactual effect of rain, $\theta(\text{rain})-\theta(\text{no rain})$, one and seven days after the onset of the event of 18 November 2024 (domain-mean rain 27\,mm on the first day); contours: 10 and 30\,mm of first-day rain.
For regional rain like this, the size of the effect is underestimated (\cref{sec:local}).}
  \label{fig:maps}
\end{figure*}

\paragraph{Responses to rain.}
Real counterfactuals are never observed, so we test the response where nature comes closest to an intervention.
A rain pulse added to the station's cell of a filtered state wets the soil as natural storms of the same size wet the probes: over 300 isolated events grouped by amount and antecedent wetness, the median increments are rank-correlated with $\rho=0.86$ in each seed, and from 10\,mm on they agree in size (0.069 and 0.128 against 0.062 and 0.108; \cref{tab:real_sm}, \cref{sec:supp_extra}); within every amount, its increment decreases with antecedent wetness.
A world model can also answer what no observation can show: removing the rain of an event from a rollout maps its effect over the domain (\cref{fig:maps}), which follows the rain field and lasts at least a week (domain mean 0.032 on day 1, 0.028 on day 7).

\begin{table*}[t]
\centering
\caption{\textbf{Local versus regional rain} (validation year unless noted).
Station soil moisture: median per-station filtering correlation and ubRMSE.
Rain response: median increment of surface soil moisture after a synthetic pulse added to the station's cell (as in \cref{tab:real_sm}) or, of the same size, to every cell of the domain; probes: isolated natural storms of the same size.
Mean $\pm$ std over three seeds; $^\ast$seed 0 only.}
\label{tab:transition}
\resizebox{\textwidth}{!}{% generated by arxiv/scripts/make_arxiv_tables.py from the JSON results -- do not edit by hand
\begin{tabular}{l ccc cc cc}
\toprule
\rowcolor{tabhead} & \multicolumn{3}{c}{Station soil moisture} & \multicolumn{2}{c}{Pulse in the station cell} & \multicolumn{2}{c}{Same pulse over the domain} \\
\cmidrule(lr){2-4}\cmidrule(lr){5-6}\cmidrule(lr){7-8}
\rowcolor{tabhead} Transition & $R$ (val.) & ubRMSE & $R$ (test) & 10--20\,mm & 20--50\,mm & 10--20\,mm & 20--50\,mm \\
\midrule
\rowcolor{tabours} \method (conserving, $3\times3$ fluxes) & 0.82$\pm$0.00 & 0.045$\pm$0.001 & 0.71$\pm$0.01 & 0.069$\pm$0.005 & 0.128$\pm$0.010 & 0.024$^\ast$ & 0.035$^\ast$ \\
\quad per-cell fluxes ($1\times1$) & 0.82$\pm$0.00 & 0.049$\pm$0.000 & 0.69$\pm$0.01 & 0.037$\pm$0.004 & 0.055$\pm$0.005 & 0.036$\pm$0.003 & 0.053$\pm$0.004 \\
\quad GRU transition & 0.81$\pm$0.01 & 0.053$\pm$0.006 & 0.68$\pm$0.00 & 0.001$\pm$0.000 & 0.002$\pm$0.000 & 0.075$\pm$0.014 & 0.112$\pm$0.020 \\
\midrule
\rowcolor{tabbase} Probes, natural storms & -- & -- & -- & 0.062 & 0.108 & 0.062 & 0.108 \\
\bottomrule
\end{tabular}
}
\end{table*}

\paragraph{Test year.}
With all settings frozen, we evaluated January--September 2025 once (18 stations report; \cref{sec:supp_test}).
All products correlate less with the probes, but the order holds: $R=0.71\pm0.01$ and anomaly $R=0.76\pm0.01$, against 0.63 and 0.66 for SMAP Level-4 (anomaly correlation higher at 13--14 of 18 stations, $p\le0.003$) and less for the LSTM and ERA5-Land.
CYGNSS alone gives 0.68 (0.63 against 0.44 for CYGNSS Level-3 on its samples), though four of seven receivers now return few valid land points (\cref{sec:supp_data}), and a pulse in the station's cell again wets the soil but barely moves that of the GRU.

\subsection{Ablations}
\paragraph{Physical decoders make the state soil moisture.}
With neural decoders (\cref{tab:real_sm,tab:real_obs}), the model forecasts the observations as well as \method (rollout NLL within 0.01 of it, two seeds), but its storage no longer follows the probes: station correlation $0.32\pm0.02$ and 10-day forecast correlation $0.31\pm0.08$.
Yet the soil water has not left its latent: a ridge probe fitted to the stations in the training period reads it ($R=0.81\pm0.01$, black box 0.81) as well as the latent of \method (0.82), whose storage needs no fit (\cref{tab:probe_real}).
Nor is the readout one of many solutions traded off by free channels and operator parameters: seeds of \method agree on the whole field (median per-cell correlation 0.99), those with neural decoders less (0.87; \cref{sec:supp_seeds}); without free channels the storage still gives $R=0.78$ (\cref{sec:supp_sens}).
As in the twin, the observation likelihood does not tell whether the state can be read without labels; the decoders do.

\paragraph{The action carries the forecast.}
Without precipitation (two seeds), rollout NLL rises to 6.02 and 2.11 and the 10-day $R$ falls to 0.63, since the observations alone must then explain each wetting of the soil.

\paragraph{The transition decides the response to local rain.}
Operators and transition divide the work: the operators make the state readable, the conserving transition makes it respond to the rain that falls on it.
A GRU transition with the same decoders tracks the probes nearly as well ($R=0.81$, ubRMSE 0.053 against 0.045), and its forecasts depend on the rain almost as much (zero-rain $\Delta$NLL 2.98 against 3.31).
Yet a storm on the station's cell hardly reaches its soil: 10--20 and 20--50\,mm pulses raise its storage by 0.001 and 0.002, against 0.062 and 0.108 at the probes, in both years (\cref{tab:real_sm,tab:test_sm}).
Without the rollout loss the model is a filter (rollout SMAP NLL 7.94 and 6.87, 10-day $R=0.70$ and 0.77) whose soil overreacts to rain (0.087 and 0.151).

\subsection{Local and regional rain}
\label{sec:local}
A pulse on one cell is an intervention outside the training distribution, since real storms span tens to hundreds of kilometres.
We therefore add the same pulse to every cell of the domain, and train a variant whose flux network sees only its own cell (per-cell fluxes, three seeds; \cref{tab:transition}).
The two transitions err in opposite directions.
The GRU, deaf to a single wet cell, responds to rain over the domain as the probes respond to natural storms (0.075 and 0.112 against 0.062 and 0.108).
\method, right for the single cell (0.062 and 0.117 in seed 0), gives rain over the domain only about a third of that (0.024 and 0.035).
Its $3\times3$ flux network explains the asymmetry: storages exchange no water, but the fraction that infiltrates depends on the rain of the neighbours, and since real rain is spatially coherent, the model has learned to send more of it to runoff when the neighbours are wet too.
Per-cell fluxes remove the asymmetry (0.037 against 0.036, and 0.055 against 0.053) but respond at about half the size of the probes, with slightly lower station skill (ubRMSE 0.049, test-year $R=0.69$).

The difference shows in forecasts.
On the 502 station-days of the validation year with at least 10\,mm of gauge rain, the soil of the GRU wets by a median $0.045\pm0.003$ over the day in 10-day rollouts, that of \method by $0.026\pm0.001$, and the probes by 0.051.
The GRU's response does come from the rain (0.001 when the rain is removed), and both correlate with the probes' change about equally ($R=0.43\pm0.02$ and $0.41\pm0.03$).
Conservation thus guarantees that every millimetre falling on a cell is booked, so the response to local rain cannot vanish; it does not guarantee that the learned split between infiltration and runoff is right.
A recurrent transition can learn the opposite shortcut, that only regional rain wets the soil, which holds in the training distribution and fails under intervention.
Judged by one kind of intervention, either model would have passed.

\section{Conclusion}
\label{sec:conclusion}
We argued that a world model of a natural system should be judged by its response to actions rather than by forecast skill, and built one for land-surface water from SMAP and CYGNSS observations.
In a twin with known counterfactuals, models trained only on observations learn the causal effect of rain, but only a physically grounded state makes this effect readable without ground truth, also when the truth contains processes the model lacks or the observations come from other physics.
Real data agree: physical decoders make the state soil moisture, which neural decoders leave readable only by a station-fitted probe; the state departs from the rain history where the satellites contradict it; and the conserving transition makes rain on a cell wet its soil as storms wet the probes, which a recurrent transition, forecasting nearly as well, does not.
The same comparison shows how much a single test can hide: under rain over the whole region the two transitions swap roles.
A learned world model should be checked under several kinds of intervention, as one would check a physical model.

\paragraph{Limitations and outlook.}
We study one region at the 5\,cm depth that probes and satellites share, treat rain as exogenous, and take the weather context from MERRA-2.
The transition has no lateral flow, and its learned split between infiltration and runoff makes the response to regional rain too weak (\cref{sec:local}); on real data, event-scale drydowns and the root zone are not yet constrained (\cref{sec:supp_dry}).
Lateral routing, observations of deeper water such as GRACE, and more varied rain during training are natural next steps.

{
    \small
    \bibliographystyle{ieeenat_fullname}
    \bibliography{main}

\begin{thebibliography}{74}
\providecommand{\natexlab}[1]{#1}
\providecommand{\url}[1]{\texttt{#1}}
\expandafter\ifx\csname urlstyle\endcsname\relax
  \providecommand{\doi}[1]{doi: #1}\else
  \providecommand{\doi}{doi: \begingroup \urlstyle{rm}\Url}\fi

\bibitem[Albughdadi(2026)]{albughdadi2026observability}
Mohanad Albughdadi.
\newblock From surface forecasting to observability forecasting: A latent world model for cloud-aware {EO} monitoring.
\newblock \emph{arXiv preprint arXiv:2607.13651}, 2026.

\bibitem[Alexe et~al.(2024)Alexe, Boucher, Lean, Pinnington, Laloyaux, McNally, Lang, Chantry, Burrows, Chrust, Pinault, Villeneuve, Bormann, and Healy]{alexe2024graphdop}
Mihai Alexe, Eulalie Boucher, Peter Lean, Ewan Pinnington, Patrick Laloyaux, Anthony McNally, Simon Lang, Matthew Chantry, Chris Burrows, Marcin Chrust, Florian Pinault, Ethel Villeneuve, Niels Bormann, and Sean Healy.
\newblock {GraphDOP}: Towards skilful data-driven medium-range weather forecasts learnt and initialised directly from observations.
\newblock \emph{arXiv preprint arXiv:2412.15687}, 2024.

\bibitem[Allen et~al.(2025)Allen, Markou, Tebbutt, et~al.]{allen2025aardvark}
Anna Allen, Stratis Markou, Will Tebbutt, et~al.
\newblock End-to-end data-driven weather prediction.
\newblock \emph{Nature}, 2025.

\bibitem[Andry et~al.(2025)Andry, Lewin, Rozet, Rochman, Mangeleer, Pirlet, Faulx, Gr{\'e}goire, and Louppe]{andry2025appa}
G{\'e}r{\^o}me Andry, Sacha Lewin, Fran{\c{c}}ois Rozet, Omer Rochman, Victor Mangeleer, Matthias Pirlet, Elise Faulx, Marilaure Gr{\'e}goire, and Gilles Louppe.
\newblock Appa: Bending weather dynamics with latent diffusion models for global data assimilation.
\newblock \emph{arXiv preprint arXiv:2504.18720}, 2025.

\bibitem[Bodnar et~al.(2025)Bodnar, Bruinsma, Lucic, et~al.]{bodnar2025aurora}
Cristian Bodnar, Wessel~P. Bruinsma, Ana Lucic, et~al.
\newblock A foundation model for the {E}arth system.
\newblock \emph{Nature}, 641:\penalty0 1180--1187, 2025.

\bibitem[Boucher et~al.(2025)Boucher, Alexe, Lean, Pinnington, Lang, Laloyaux, Zampieri, de~Rosnay, Bormann, and McNally]{boucher2025coupled}
Eulalie Boucher, Mihai Alexe, Peter Lean, Ewan Pinnington, Simon Lang, Patrick Laloyaux, Lorenzo Zampieri, Patricia de Rosnay, Niels Bormann, and Anthony McNally.
\newblock Learning coupled {E}arth system dynamics with {GraphDOP}.
\newblock \emph{arXiv preprint arXiv:2510.20416}, 2025.

\bibitem[Brocca et~al.(2014)Brocca, Ciabatta, Massari, Moramarco, Hahn, Hasenauer, Kidd, Dorigo, Wagner, and Levizzani]{brocca2014sm2rain}
Luca Brocca, Luca Ciabatta, Christian Massari, Tommaso Moramarco, Sebastian Hahn, Stefan Hasenauer, Richard Kidd, Wouter Dorigo, Wolfgang Wagner, and Vincenzo Levizzani.
\newblock Soil as a natural rain gauge: Estimating global rainfall from satellite soil moisture data.
\newblock \emph{Journal of Geophysical Research: Atmospheres}, 119\penalty0 (9):\penalty0 5128--5141, 2014.

\bibitem[Brodzik et~al.(2012)Brodzik, Billingsley, Haran, Raup, and Savoie]{brodzik2012ease2}
Mary~J. Brodzik, Brendan Billingsley, Terry Haran, Bruce Raup, and Matthew~H. Savoie.
\newblock {EASE-Grid} 2.0: Incremental but significant improvements for earth-gridded data sets.
\newblock \emph{ISPRS International Journal of Geo-Information}, 1\penalty0 (1):\penalty0 32--45, 2012.

\bibitem[Chen et~al.(2026)Chen, Qi, Zhang, Jin, Zhang, and Liu]{chen2026gsreasoner}
Yiming Chen, Zekun Qi, Wenyao Zhang, Xin Jin, Li Zhang, and Peidong Liu.
\newblock Reasoning in space via grounding in the world.
\newblock In \emph{International Conference on Learning Representations (ICLR)}, 2026.

\bibitem[Chew et~al.(2018)Chew, Reager, and Small]{chew2018inundation}
Clara Chew, John~T. Reager, and Eric Small.
\newblock {CYGNSS} data map flood inundation during the 2017 {A}tlantic hurricane season.
\newblock \emph{Scientific Reports}, 8:\penalty0 9336, 2018.

\bibitem[Chew and Small(2018)]{chew2018smap}
Clara~C. Chew and Eric~E. Small.
\newblock Soil moisture sensing using spaceborne {GNSS} reflections: Comparison of {CYGNSS} reflectivity to {SMAP} soil moisture.
\newblock \emph{Geophysical Research Letters}, 45\penalty0 (9):\penalty0 4049--4057, 2018.

\bibitem[Crow et~al.(2011)Crow, van~den Berg, Huffman, and Pellarin]{crow2011smart}
Wade~T. Crow, Martinus~J. van~den Berg, George~J. Huffman, and Thierry Pellarin.
\newblock Correcting rainfall using satellite-based surface soil moisture retrievals: The {Soil Moisture Analysis Rainfall Tool (SMART)}.
\newblock \emph{Water Resources Research}, 47\penalty0 (8), 2011.

\bibitem[De~Roo and Ulaby(1994)]{deroo1994bistatic}
Roger~D. De~Roo and Fawwaz~T. Ulaby.
\newblock Bistatic specular scattering from rough dielectric surfaces.
\newblock \emph{IEEE Transactions on Antennas and Propagation}, 42\penalty0 (2):\penalty0 220--231, 1994.

\bibitem[Diamond et~al.(2013)Diamond, Karl, Palecki, et~al.]{diamond2013uscrn}
Howard~J. Diamond, Thomas~R. Karl, Michael~A. Palecki, et~al.
\newblock {U.S.} climate reference network after one decade of operations: Status and assessment.
\newblock \emph{Bulletin of the American Meteorological Society}, 94\penalty0 (4):\penalty0 485--498, 2013.

\bibitem[Entekhabi et~al.(2010)Entekhabi, Njoku, O'Neill, et~al.]{entekhabi2010smap}
Dara Entekhabi, Eni~G. Njoku, Peggy~E. O'Neill, et~al.
\newblock The soil moisture active passive ({SMAP}) mission.
\newblock \emph{Proceedings of the IEEE}, 98\penalty0 (5):\penalty0 704--716, 2010.

\bibitem[Eroglu et~al.(2019)Eroglu, Kurum, Boyd, and Gurbuz]{eroglu2019cygnss}
Orhan Eroglu, Mehmet Kurum, Dylan Boyd, and Ali~Cafer Gurbuz.
\newblock High spatio-temporal resolution {CYGNSS} soil moisture estimates using artificial neural networks.
\newblock \emph{Remote Sensing}, 11\penalty0 (19):\penalty0 2272, 2019.

\bibitem[Fang and Shen(2020)]{fang2020nrt}
Kuai Fang and Chaopeng Shen.
\newblock Near-real-time forecast of satellite-based soil moisture using long short-term memory with an adaptive data integration kernel.
\newblock \emph{Journal of Hydrometeorology}, 21\penalty0 (3):\penalty0 399--413, 2020.

\bibitem[Fang et~al.(2017)Fang, Shen, Kifer, and Yang]{fang2017prolongation}
Kuai Fang, Chaopeng Shen, Daniel Kifer, and Xiao Yang.
\newblock Prolongation of {SMAP} to spatiotemporally seamless coverage of continental {U.S.} using a deep learning neural network.
\newblock \emph{Geophysical Research Letters}, 44\penalty0 (21):\penalty0 11030--11039, 2017.

\bibitem[Feng et~al.(2022)Feng, Liu, Lawson, and Shen]{feng2022differentiable}
Dapeng Feng, Jiangtao Liu, Kathryn Lawson, and Chaopeng Shen.
\newblock Differentiable, learnable, regionalized process-based models with multiphysical outputs can approach state-of-the-art hydrologic prediction accuracy.
\newblock \emph{Water Resources Research}, 58\penalty0 (10):\penalty0 e2022WR032404, 2022.

\bibitem[Feng et~al.(2026)Feng, Li, Zhou, Butterbach-Bahl, Guan, Jin, Looms, Wang, Igel, Treat, Olesen, and Wang]{feng2026pgml}
Sijia Feng, Aoyang Li, Rui Zhou, Klaus Butterbach-Bahl, Kaiyu Guan, Zhenong Jin, Majken~C. Looms, Sherrie Wang, Christian Igel, Claire Treat, J{\o}rgen~Eivind Olesen, and Sheng Wang.
\newblock Global daily 9 km remotely sensed soil moisture (2015--2025) with microwave radiative transfer-guided learning.
\newblock \emph{Scientific Data}, 13:\penalty0 435, 2026.

\bibitem[Gelaro et~al.(2017)Gelaro, McCarty, Su{\'a}rez, et~al.]{gelaro2017merra2}
Ronald Gelaro, Will McCarty, Max~J. Su{\'a}rez, et~al.
\newblock The modern-era retrospective analysis for research and applications, version 2 ({MERRA-2}).
\newblock \emph{Journal of Climate}, 30\penalty0 (14):\penalty0 5419--5454, 2017.

\bibitem[Gong et~al.(2026)Gong, Xu, Wei, Tu, Xu, Liu, et~al.]{gong2026eartho1}
Junchao Gong, Kaiyi Xu, Wangxu Wei, Siwei Tu, Jingyi Xu, Zili Liu, et~al.
\newblock {Earth-o1}: A grid-free observation-native atmospheric world model.
\newblock \emph{arXiv preprint arXiv:2605.06337}, 2026.

\bibitem[Gu et~al.(2022)Gu, Goel, and R{\'e}]{gu2022s4}
Albert Gu, Karan Goel, and Christopher R{\'e}.
\newblock Efficiently modeling long sequences with structured state spaces.
\newblock In \emph{International Conference on Learning Representations}, 2022.

\bibitem[Ha and Schmidhuber(2018)]{ha2018world}
David Ha and J{\"u}rgen Schmidhuber.
\newblock Recurrent world models facilitate policy evolution.
\newblock In \emph{Advances in Neural Information Processing Systems}, 2018.

\bibitem[Hafner et~al.(2019)Hafner, Lillicrap, Fischer, Villegas, Ha, Lee, and Davidson]{hafner2019planet}
Danijar Hafner, Timothy Lillicrap, Ian Fischer, Ruben Villegas, David Ha, Honglak Lee, and James Davidson.
\newblock Learning latent dynamics for planning from pixels.
\newblock In \emph{International Conference on Machine Learning}, 2019.

\bibitem[Hafner et~al.(2025)Hafner, Pasukonis, Ba, and Lillicrap]{hafner2025dreamerv3}
Danijar Hafner, Jurgis Pasukonis, Jimmy Ba, and Timothy Lillicrap.
\newblock Mastering diverse control tasks through world models.
\newblock \emph{Nature}, 640:\penalty0 647--653, 2025.

\bibitem[Hakim and Masanam(2023)]{hakim2023dynamical}
Gregory~J. Hakim and Sanjit Masanam.
\newblock Dynamical tests of a deep-learning weather prediction model.
\newblock \emph{arXiv preprint arXiv:2309.10867}, 2023.

\bibitem[Hoedt et~al.(2021)Hoedt, Kratzert, Klotz, Halmich, Holzleitner, Nearing, Hochreiter, and Klambauer]{hoedt2021mclstm}
Pieter-Jan Hoedt, Frederik Kratzert, Daniel Klotz, Christina Halmich, Markus Holzleitner, Grey Nearing, Sepp Hochreiter, and G{\"u}nter Klambauer.
\newblock {MC-LSTM}: Mass-conserving {LSTM}.
\newblock In \emph{International Conference on Machine Learning (ICML)}, pages 4275--4286, 2021.

\bibitem[Huffman et~al.(2020)Huffman, Bolvin, Braithwaite, et~al.]{huffman2020imerg}
George~J. Huffman, David~T. Bolvin, Dan Braithwaite, et~al.
\newblock Integrated multi-satellite retrievals for the global precipitation measurement ({GPM}) mission ({IMERG}).
\newblock In \emph{Satellite Precipitation Measurement}, pages 343--353. Springer, 2020.

\bibitem[Iele et~al.(2026)Iele, Mulero~Ayll{\'o}n, Soda, and Tortora]{iele2026vegsim}
Irene Iele, Elena Mulero~Ayll{\'o}n, Paolo Soda, and Matteo Tortora.
\newblock {VegSim}: A geospatial world model for scenario-conditioned vegetation simulation.
\newblock \emph{arXiv preprint arXiv:2606.21961}, 2026.

\bibitem[Jamaat et~al.(2025)Jamaat, Song, Rahmani, Liu, Lawson, and Shen]{jamaat2025update}
Amirmoez Jamaat, Yalan Song, Farshid Rahmani, Jiangtao Liu, Kathryn Lawson, and Chaopeng Shen.
\newblock Update hydrological states or meteorological forcings? comparing data assimilation methods for differentiable hydrologic models.
\newblock \emph{arXiv preprint arXiv:2502.16444}, 2025.

\bibitem[Jia et~al.(2026)Jia, Qi, Zhang, Zhang, Yu, He, Wang, and Yi]{jia2026omnispatial}
Mengdi Jia, Zekun Qi, Shaochen Zhang, Wenyao Zhang, Xinqiang Yu, Jiawei He, He Wang, and Li Yi.
\newblock {OmniSpatial}: Towards comprehensive spatial reasoning benchmark for vision language models.
\newblock In \emph{International Conference on Learning Representations (ICLR)}, 2026.

\bibitem[Kang et~al.(2024)Kang, Yue, Lu, Lin, Zhao, Wang, Huang, and Feng]{kang2024howfar}
Bingyi Kang, Yang Yue, Rui Lu, Zhijie Lin, Yang Zhao, Kaixin Wang, Gao Huang, and Jiashi Feng.
\newblock How far is video generation from world model: A physical law perspective.
\newblock \emph{arXiv preprint arXiv:2411.02385}, 2024.

\bibitem[Khemakhem et~al.(2020)Khemakhem, Kingma, Monti, and Hyv{\"a}rinen]{khemakhem2020ivae}
Ilyes Khemakhem, Diederik~P. Kingma, Ricardo~Pio Monti, and Aapo Hyv{\"a}rinen.
\newblock Variational autoencoders and nonlinear {ICA}: A unifying framework.
\newblock In \emph{Proceedings of the Twenty Third International Conference on Artificial Intelligence and Statistics (AISTATS)}, pages 2207--2217, 2020.

\bibitem[Koster et~al.(2004)Koster, Dirmeyer, Guo, et~al.]{koster2004regions}
Randal~D. Koster, Paul~A. Dirmeyer, Zhichang Guo, et~al.
\newblock Regions of strong coupling between soil moisture and precipitation.
\newblock \emph{Science}, 305\penalty0 (5687):\penalty0 1138--1140, 2004.

\bibitem[Kratzert et~al.(2019)Kratzert, Klotz, Shalev, Klambauer, Hochreiter, and Nearing]{kratzert2019towards}
Frederik Kratzert, Daniel Klotz, Guy Shalev, G{\"u}nter Klambauer, Sepp Hochreiter, and Grey Nearing.
\newblock Towards learning universal, regional, and local hydrological behaviors via machine learning applied to large-sample datasets.
\newblock \emph{Hydrology and Earth System Sciences}, 23:\penalty0 5089--5110, 2019.

\bibitem[Krishnan et~al.(2017)Krishnan, Shalit, and Sontag]{krishnan2017structured}
Rahul~G. Krishnan, Uri Shalit, and David Sontag.
\newblock Structured inference networks for nonlinear state space models.
\newblock In \emph{AAAI Conference on Artificial Intelligence}, 2017.

\bibitem[Lam et~al.(2023)Lam, Sanchez-Gonzalez, Willson, et~al.]{lam2023graphcast}
Remi Lam, Alvaro Sanchez-Gonzalez, Matthew Willson, et~al.
\newblock Learning skillful medium-range global weather forecasting.
\newblock \emph{Science}, 382\penalty0 (6677):\penalty0 1416--1421, 2023.

\bibitem[Lean et~al.(2025)Lean, Alexe, Boucher, Pinnington, Lang, Laloyaux, Bormann, and McNally]{lean2025graphdoprep}
Peter Lean, Mihai Alexe, Eulalie Boucher, Ewan Pinnington, Simon Lang, Patrick Laloyaux, Niels Bormann, and Anthony McNally.
\newblock Learning from nature: Insights into {GraphDOP}'s representations of the {E}arth system.
\newblock \emph{arXiv preprint arXiv:2508.18018}, 2025.

\bibitem[Locatello et~al.(2019)Locatello, Bauer, Lucic, R{\"a}tsch, Gelly, Sch{\"o}lkopf, and Bachem]{locatello2019challenging}
Francesco Locatello, Stefan Bauer, Mario Lucic, Gunnar R{\"a}tsch, Sylvain Gelly, Bernhard Sch{\"o}lkopf, and Olivier Bachem.
\newblock Challenging common assumptions in the unsupervised learning of disentangled representations.
\newblock In \emph{Proceedings of the 36th International Conference on Machine Learning (ICML)}, 2019.

\bibitem[Luo et~al.(2026)Luo, Yuan, Yang, Li, Liu, and Zhao]{luo2026eowm}
Junwei Luo, Shuai Yuan, Zhenya Yang, Yansheng Li, Zhe Liu, and Hengshuang Zhao.
\newblock {EO-WM}: A physically informed world model for probabilistic earth observation forecasting.
\newblock \emph{arXiv preprint arXiv:2606.27277}, 2026.

\bibitem[McColl et~al.(2017)McColl, Alemohammad, Akbar, Konings, Yueh, and Entekhabi]{mccoll2017global}
Kaighin~A. McColl, Seyed~Hamed Alemohammad, Ruzbeh Akbar, Alexandra~G. Konings, Simon Yueh, and Dara Entekhabi.
\newblock The global distribution and dynamics of surface soil moisture.
\newblock \emph{Nature Geoscience}, 10:\penalty0 100--104, 2017.

\bibitem[Mironov et~al.(2009)Mironov, Kosolapova, and Fomin]{mironov2009dielectric}
Valery~L. Mironov, Liudmila~G. Kosolapova, and Sergey~V. Fomin.
\newblock Physically and mineralogically based spectroscopic dielectric model for moist soils.
\newblock \emph{IEEE Transactions on Geoscience and Remote Sensing}, 47\penalty0 (7):\penalty0 2059--2070, 2009.

\bibitem[Mo et~al.(1982)Mo, Choudhury, Schmugge, Wang, and Jackson]{mo1982tauomega}
T. Mo, B.~J. Choudhury, T.~J. Schmugge, J.~R. Wang, and T.~J. Jackson.
\newblock A model for microwave emission from vegetation-covered fields.
\newblock \emph{Journal of Geophysical Research: Oceans}, 87\penalty0 (C13):\penalty0 11229--11237, 1982.

\bibitem[Morris et~al.(2019)Morris, Chew, Reager, Shah, and Zuffada]{morris2019everglades}
Mary Morris, Clara Chew, John~T. Reager, Rashmi Shah, and Cinzia Zuffada.
\newblock A novel approach to monitoring wetland dynamics using {CYGNSS}: {E}verglades case study.
\newblock \emph{Remote Sensing of Environment}, 233:\penalty0 111417, 2019.

\bibitem[Motamed et~al.(2025)Motamed, Culp, Swersky, Jaini, and Geirhos]{motamed2025physicsiq}
Saman Motamed, Laura Culp, Kevin Swersky, Priyank Jaini, and Robert Geirhos.
\newblock Do generative video models understand physical principles?
\newblock \emph{arXiv preprint arXiv:2501.09038}, 2025.

\bibitem[Mu{\~n}oz-Sabater et~al.(2021)Mu{\~n}oz-Sabater, Dutra, Agust{\'\i}-Panareda, et~al.]{munozsabater2021era5land}
Joaqu{\'\i}n Mu{\~n}oz-Sabater, Emanuel Dutra, Anna Agust{\'\i}-Panareda, et~al.
\newblock {ERA5-Land}: A state-of-the-art global reanalysis dataset for land applications.
\newblock \emph{Earth System Science Data}, 13:\penalty0 4349--4383, 2021.

\bibitem[Nearing et~al.(2024)Nearing, Cohen, Dube, et~al.]{nearing2024floods}
Grey Nearing, Deborah Cohen, Vusumuzi Dube, et~al.
\newblock Global prediction of extreme floods in ungauged watersheds.
\newblock \emph{Nature}, 627:\penalty0 559--563, 2024.

\bibitem[{NVIDIA} et~al.(2025){NVIDIA}, Agarwal, Ali, et~al.]{nvidia2025cosmos}
{NVIDIA}, Niket Agarwal, Arslan Ali, et~al.
\newblock Cosmos world foundation model platform for physical {AI}.
\newblock \emph{arXiv preprint arXiv:2501.03575}, 2025.

\bibitem[Orvieto et~al.(2023)Orvieto, Smith, Gu, Fernando, Gulcehre, Pascanu, and De]{orvieto2023lru}
Antonio Orvieto, Samuel~L. Smith, Albert Gu, Anushan Fernando, Caglar Gulcehre, Razvan Pascanu, and Soham De.
\newblock Resurrecting recurrent neural networks for long sequences.
\newblock In \emph{International Conference on Machine Learning}, 2023.

\bibitem[Peng et~al.(2026)Peng, Zhang, Xu, Li, Zhu, Qi, Liu, Zhang, Zeng, and Jin]{peng2026reworld}
Baorui Peng, Wenyao Zhang, Liang Xu, Bohan Li, Hu Zhu, Zekun Qi, Hongsi Liu, Jiazhao Zhang, Wenjun Zeng, and Xin Jin.
\newblock {ReWorld}: Multi-dimensional reward modeling for embodied video generation.
\newblock \emph{arXiv preprint arXiv:2601.12428}, 2026.

\bibitem[Price et~al.(2025)Price, Sanchez-Gonzalez, Alet, et~al.]{price2025gencast}
Ilan Price, Alvaro Sanchez-Gonzalez, Ferran Alet, et~al.
\newblock Probabilistic weather forecasting with machine learning.
\newblock \emph{Nature}, 637:\penalty0 84--90, 2025.

\bibitem[Qi et~al.(2024)Qi, Dong, Zhang, Geng, Han, Ge, Yi, and Ma]{qi2024shapellm}
Zekun Qi, Runpei Dong, Shaochen Zhang, Haoran Geng, Chunrui Han, Zheng Ge, Li Yi, and Kaisheng Ma.
\newblock {ShapeLLM}: Universal {3D} object understanding for embodied interaction.
\newblock In \emph{European Conference on Computer Vision (ECCV)}, 2024.

\bibitem[Raoult et~al.(2026)Raoult, Pinnington, Santa~Cruz, Pinault, Raoult, Zelenka, Arduini, Balsamo, Boussetta, Chantry, de~Rosnay, Dueben, and R{\"u}diger]{raoult2026ailand}
Nina Raoult, Ewan Pinnington, Mario Santa~Cruz, Florian Pinault, Baudouin Raoult, Natalie Zelenka, Gabriele Arduini, Gianpaolo Balsamo, Souhail Boussetta, Matthew Chantry, Patricia de Rosnay, Peter Dueben, and Christoph R{\"u}diger.
\newblock {aiLand} v1: Physics-based land surface emulator with observational fine-tuning.
\newblock \emph{EGUsphere preprint}, 2026.

\bibitem[Reichle et~al.(2017)Reichle, De~Lannoy, Liu, et~al.]{reichle2017smapl4}
Rolf~H. Reichle, Gabri{\"e}lle J.~M. De~Lannoy, Qing Liu, et~al.
\newblock Global assessment of the {SMAP} level-4 surface and root-zone soil moisture product using assimilation diagnostics.
\newblock \emph{Journal of Hydrometeorology}, 18\penalty0 (12):\penalty0 3217--3237, 2017.

\bibitem[Rozet and Louppe(2023)]{rozet2023sda}
Fran{\c{c}}ois Rozet and Gilles Louppe.
\newblock Score-based data assimilation.
\newblock In \emph{Advances in Neural Information Processing Systems}, 2023.

\bibitem[Rubanova et~al.(2019)Rubanova, Chen, and Duvenaud]{rubanova2019latentode}
Yulia Rubanova, Ricky T.~Q. Chen, and David Duvenaud.
\newblock Latent ordinary differential equations for irregularly-sampled time series.
\newblock In \emph{Advances in Neural Information Processing Systems}, 2019.

\bibitem[Ruf et~al.(2016)Ruf, Atlas, Chang, et~al.]{ruf2016cygnss}
Christopher~S. Ruf, Robert Atlas, Paul~S. Chang, et~al.
\newblock New ocean winds satellite mission to probe hurricanes and tropical convection.
\newblock \emph{Bulletin of the American Meteorological Society}, 97\penalty0 (3):\penalty0 385--395, 2016.

\bibitem[Schaefer et~al.(2007)Schaefer, Cosh, and Jackson]{schaefer2007scan}
Garry~L. Schaefer, Michael~H. Cosh, and Thomas~J. Jackson.
\newblock The {USDA} natural resources conservation service soil climate analysis network ({SCAN}).
\newblock \emph{Journal of Atmospheric and Oceanic Technology}, 24\penalty0 (12):\penalty0 2073--2077, 2007.

\bibitem[Shen et~al.(2023)Shen, Appling, Gentine, et~al.]{shen2023differentiable}
Chaopeng Shen, Alison~P. Appling, Pierre Gentine, et~al.
\newblock Differentiable modelling to unify machine learning and physical models for geosciences.
\newblock \emph{Nature Reviews Earth \& Environment}, 4:\penalty0 552--567, 2023.

\bibitem[Si and Chen(2025)]{si2025latentensf}
Phillip Si and Peng Chen.
\newblock {Latent-EnSF}: A latent ensemble score filter for high-dimensional data assimilation with sparse observation data.
\newblock In \emph{International Conference on Learning Representations}, 2025.

\bibitem[Sun et~al.(2026)Sun, Zhang, Qi, Ren, Liu, Zhu, Sun, Jin, and Chen]{sun2026vlajepa}
Jingwen Sun, Wenyao Zhang, Zekun Qi, Shaojie Ren, Zezhi Liu, Hanxin Zhu, Guangzhong Sun, Xin Jin, and Zhibo Chen.
\newblock {VLA-JEPA}: Enhancing vision-language-action model with latent world model.
\newblock In \emph{European Conference on Computer Vision (ECCV)}, 2026.

\bibitem[Takeishi and Kalousis(2021)]{takeishi2021pivae}
Naoya Takeishi and Alexandros Kalousis.
\newblock Physics-integrated variational autoencoders for robust and interpretable generative modeling.
\newblock In \emph{Advances in Neural Information Processing Systems (NeurIPS)}, 2021.

\bibitem[Tr{\'e}molet(2006)]{tremolet2006weak}
Yannick Tr{\'e}molet.
\newblock Accounting for an imperfect model in {4D-Var}.
\newblock \emph{Quarterly Journal of the Royal Meteorological Society}, 132\penalty0 (621):\penalty0 2483--2504, 2006.

\bibitem[Unwin et~al.(2021)Unwin, Pierdicca, Cardellach, et~al.]{unwin2021hydrognss}
Martin~J. Unwin, Nazzareno Pierdicca, Estel Cardellach, et~al.
\newblock An introduction to the {HydroGNSS} {GNSS} reflectometry remote sensing mission.
\newblock \emph{IEEE Journal of Selected Topics in Applied Earth Observations and Remote Sensing}, 14:\penalty0 6987--6999, 2021.

\bibitem[Vafa et~al.(2024)Vafa, Chen, Rambachan, Kleinberg, and Mullainathan]{vafa2024evaluating}
Keyon Vafa, Justin~Y. Chen, Ashesh Rambachan, Jon Kleinberg, and Sendhil Mullainathan.
\newblock Evaluating the world model implicit in a generative model.
\newblock \emph{arXiv preprint arXiv:2406.03689}, 2024.

\bibitem[Vafa et~al.(2025)Vafa, Chang, Rambachan, and Mullainathan]{vafa2025found}
Keyon Vafa, Peter~G. Chang, Ashesh Rambachan, and Sendhil Mullainathan.
\newblock What has a foundation model found? using inductive bias to probe for world models.
\newblock \emph{arXiv preprint arXiv:2507.06952}, 2025.

\bibitem[van Genuchten(1980)]{vangenuchten1980closed}
M.~Th. van Genuchten.
\newblock A closed-form equation for predicting the hydraulic conductivity of unsaturated soils.
\newblock \emph{Soil Science Society of America Journal}, 44\penalty0 (5):\penalty0 892--898, 1980.

\bibitem[Vural et~al.(2026)Vural, Vanderbecken, Bonan, Rojas-Munoz, and Calvet]{vural2026nnoperator}
Jasmin Vural, Pierre Vanderbecken, Bertrand Bonan, Oscar Rojas-Munoz, and Jean-Christophe Calvet.
\newblock A neural-network based observation operator for the assimilation of microwave satellite observations with {LDAS-Monde} on global cropland for improving land-surface variables.
\newblock \emph{Earth Observation}, 1:\penalty0 155--176, 2026.

\bibitem[Wang et~al.(2026)Wang, Wang, Li, Ma, Hurtt, Jia, Mai, Wang, and Xie]{wang2026whatif}
Zhihao Wang, Ruichen Wang, Ruohan Li, Lei Ma, George Hurtt, Xiaowei Jia, Gengchen Mai, Shaowen Wang, and Yiqun Xie.
\newblock Earth system world model for what-if simulations: A case study for terrestrial ecosystems.
\newblock \emph{arXiv preprint arXiv:2609.08855}, 2026.

\bibitem[Yin et~al.(2021)Yin, Le~Guen, Dona, de~B{\'e}zenac, Ayed, Thome, and Gallinari]{yin2021aphynity}
Yuan Yin, Vincent Le~Guen, J{\'e}r{\'e}mie Dona, Emmanuel de B{\'e}zenac, Ibrahim Ayed, Nicolas Thome, and Patrick Gallinari.
\newblock Augmenting physical models with deep networks for complex dynamics forecasting.
\newblock In \emph{International Conference on Learning Representations (ICLR)}, 2021.

\bibitem[Zhang et~al.(2025)Zhang, Liu, Qi, Wang, Yu, Zhang, Dong, He, Lu, Wang, Zhang, Yi, Zeng, and Jin]{zhang2025dreamvla}
Wenyao Zhang, Hongsi Liu, Zekun Qi, Yunnan Wang, Xinqiang Yu, Jiazhao Zhang, Runpei Dong, Jiawei He, Fan Lu, He Wang, Zhizheng Zhang, Li Yi, Wenjun Zeng, and Xin Jin.
\newblock {DreamVLA}: A vision-language-action model dreamed with comprehensive world knowledge.
\newblock In \emph{Advances in Neural Information Processing Systems (NeurIPS)}, 2025.

\bibitem[Zhang et~al.(2026{\natexlab{a}})Zhang, Zhang, Qi, Zeng, Jin, and Zhang]{zhang2026defi}
Wenyao Zhang, Bozhou Zhang, Zekun Qi, Wenjun Zeng, Xin Jin, and Li Zhang.
\newblock Disentangled robot learning via separate forward and inverse dynamics pretraining.
\newblock In \emph{International Conference on Learning Representations (ICLR)}, 2026{\natexlab{a}}.

\bibitem[Zhang et~al.(2026{\natexlab{b}})Zhang, Zhang, Qi, Zhang, Lin, Zhang, Mu, Yang, Zeng, and Jin]{zhang2026imagewam}
Yuyang Zhang, Wenyao Zhang, Zekun Qi, He Zhang, Haitao Lin, Jingbo Zhang, Yao Mu, Xiaokang Yang, Wenjun Zeng, and Xin Jin.
\newblock {ImageWAM}: Do world action models really need video generation, or just image editing?
\newblock In \emph{Conference on Robot Learning (CoRL)}, 2026{\natexlab{b}}.

\end{thebibliography}
}

\clearpage
\maketitlesupplementary
\setcounter{section}{0}\setcounter{table}{0}\setcounter{figure}{0}\setcounter{equation}{0}
\renewcommand{\thesection}{S\arabic{section}}\renewcommand{\thetable}{S\arabic{table}}\renewcommand{\thefigure}{S\arabic{figure}}\renewcommand{\theequation}{S\arabic{equation}}
\renewcommand{\theHsection}{S\arabic{section}}\renewcommand{\theHtable}{S\arabic{table}}\renewcommand{\theHfigure}{S\arabic{figure}}\renewcommand{\theHequation}{S\arabic{equation}}

\section{Data sources and preprocessing}
\label{sec:supp_data}
\cref{tab:supp_data} lists all inputs. All data are openly available.
Granules are subset on the server through NASA's cloud OPeNDAP service (CYGNSS, IMERG, MERRA-2) or, for SMAP, read remotely by byte ranges; only the study window is kept, and no full granules are stored.
For the baselines, SMAP Level-4 (SPL4SMGP v008) is subset through OPeNDAP on the same 9\,km grid, and ERA5-Land 0--7\,cm soil moisture is taken at the grid point nearest each station (median distance 3.8\,km) from the Open-Meteo historical archive, which redistributes ERA5-Land~\cite{munozsabater2021era5land}.
On a sample of granules, both access paths give bit-identical extracts.
For CYGNSS, only per-specular-point scalars are read; the delay--Doppler maps are never read.
CYGNSS passes over the region are clustered in a window of a few hours each day, and the window drifts from day to day with the orbit.
The number of CYGNSS tokens per 6\,h step is therefore very uneven.
Static layers are area-averaged onto the 9\,km EASE-Grid~2.0 window ($95\times108$ cells).
Elevation and slope are computed on a mosaic of all elevation tiles before averaging, which avoids tile-boundary artefacts.
\begin{table*}[t]
\centering\small
\resizebox{\textwidth}{!}{%
\begin{tabular}{llll}
\toprule
\rowcolor{tabhead} Role & Product & Native resolution & Use \\
\midrule
Observation & SMAP enhanced L1C brightness temperature (SPL1CTB\_E) & 9\,km EASE-2, $\sim$2--3\,d revisit & H/V, fore and aft looks as separate tokens \\
Observation & CYGNSS Level-1 v3.2 & specular points & peak reflectivity, geometry, quality flags \\
Action & GPM IMERG Final V07~\cite{huffman2020imerg} & 0.1$^\circ$, 30\,min & summed to 6\,h \\
Context & MERRA-2~\cite{gelaro2017merra2} (single-level and radiation) & 0.5$^\circ\times$0.625$^\circ$, 1\,h & 6\,h means; forcing variables only \\
Static & Copernicus DEM GLO-90 & 90\,m & mean elevation, sub-grid relief, slope \\
Static & SoilGrids 2.0 & 250\,m & clay, sand (0--5, 5--15\,cm), organic carbon \\
Static & ESA WorldCover 2021 v200 & 10\,m & class fractions \\
Static & JRC Global Surface Water v1.4 & 30\,m & water occurrence \\
Evaluation & USCRN~\cite{diamond2013uscrn}, SCAN~\cite{schaefer2007scan} & stations, hourly & 5\,cm soil moisture, precipitation \\
\bottomrule
\end{tabular}}
\caption{\textbf{Data used in this work.} Evaluation data are never used for training or model selection.}
\label{tab:supp_data}
\end{table*}

\paragraph{Quality control.}
\begin{itemize}
\item \textbf{SMAP}: tokens require finite brightness temperatures in $[100,330]$\,K and a clear quality bit.
\item \textbf{CYGNSS}: specular points require all of the following:
  \begin{itemize}
  \item incidence $\le65^\circ$;
  \item receive-antenna gain $\ge0$\,dBi;
  \item delay--Doppler SNR $\ge2$\,dB;
  \item a valid land flag;
  \item the \texttt{poor\_overall\_quality} bit not set (resolved from the product's own flag metadata).
  \end{itemize}
\item \textbf{Assignment}: each observation is assigned to the nearest 6\,h state time; its offset in hours is a token feature.
\item \textbf{In-situ data}: SCAN values are kept only when flagged valid; local standard times are converted to UTC; range, frozen-soil and spike checks are applied; station values are averaged over the 12\,h around each state time.
\end{itemize}
After quality control, August 2018--September 2025 holds 46.9\,M SMAP tokens (45\% of the 6\,h steps observed; every one of the 10{,}260 cells observed) and 23.9\,M CYGNSS tokens (53\% of the steps; 68 cells never observed).
CYGNSS tokens per year peak in 2020 (4.6\,M) and fall to 2.5\,M in 2024.
The delivered specular points do not decrease, except for FM06, which has returned no data since November 2022.
The decrease comes from the land flag: the Level-1 processor marks a growing share of land specular points of FM01, FM05 and FM07 (from 2024 also FM08) as invalid, down to 2--4\% valid in 2025 against 95--97\% for FM02--FM04.
We keep this flag, since a wrong specular-point location would place the observation in the wrong cell, so the 2025 test year is observed mostly by FM02--FM04 and FM08.
The CYGNSS decoder has a bias embedding per receiver--transmitter pair (\cref{sec:decoders}), which absorbs differences between instruments.

\paragraph{Context variables.}
The context consists of 2\,m temperature, 2\,m specific humidity, 10\,m wind speed, surface pressure, downward shortwave and absorbed longwave radiation, and the sine and cosine of the hour of day.
They are normalized with fixed physical constants rather than data statistics.
We deliberately exclude land-model outputs (skin temperature, albedo), which would leak the land state.
We also exclude calendar features, which would let the model explain seasonality without the action.

\section{Twin experiment details}
\label{sec:supp_twin}
It has two soil layers (50\,mm and 950\,mm), power-law infiltration, critical-moisture limitation of evapotranspiration, nonlinear drainage, and threshold ponding controlled by height above drainage.
Its parameters (porosity, critical moisture, saturated conductivity, roughness, vegetation amplitude) are smooth random fields that the model never sees directly.
The model only receives noisy proxies of texture, terrain and aridity as static inputs.
Precipitation comes from seasonally modulated Gaussian storm cells.
The storm rate is proportional to domain area, and the west--east gradient is calibrated to a Southern-Great-Plains-like climatology (about 460 to 1120\,mm\,yr$^{-1}$ on the $64\times64$ domain).
Context fields mimic the meteorological forcing of the real setting: skin temperature, potential evaporation, shortwave-like and humidity-like fields, and hour of day.
As in the real setting, the context has no calendar features, so seasonality reaches the model only through forcing.
SMAP-like overpasses are at 06/18\,h on swaths covering 45\% of the domain; CYGNSS-like specular points have random incidence angles, per-receiver/transmitter biases of 0.7\,dB and open-water mixing from ponded storage.

\begin{table*}[t]
\centering
\caption{\textbf{Twin experiment: prediction, readability and information.}
Direct $R$: designated surface-storage channel against true surface soil moisture, no fitting.
Probe $R$: a ridge probe fit on the training period.
Rollout NLL: per-token NLL of observations in 10-day open-loop rollouts; zero-rain: rain removed.
Mean $\pm$ std over seeds and worlds (\method, no residual: three; others: two or one).}
\label{tab:twin_full}
\resizebox{\textwidth}{!}{% generated by code/scripts/make_twin_table.py -- do not edit by hand
\begin{tabular}{l cc cc c cc c}
\toprule
\rowcolor{tabhead}  & \multicolumn{2}{c}{Direct $R$} & \multicolumn{2}{c}{Probe $R$} & Assim. & \multicolumn{2}{c}{Rollout NLL} & Zero-rain \\
\cmidrule(lr){2-3}\cmidrule(lr){4-5}\cmidrule(lr){7-8}
\rowcolor{tabhead} Model & filter & rollout & filter & rollout & gain $\Delta R$ & SMAP & CYGNSS & SMAP NLL \\
\midrule
\rowcolor{tabours} \method & 0.78$\pm$0.03 & 0.83$\pm$0.03 & 0.79$\pm$0.03 & 0.84$\pm$0.03 & 0.015$\pm$0.015 & 3.90$\pm$0.02 & 1.47$\pm$0.01 & 10.55$\pm$0.34 \\
\quad no residual content & 0.78$\pm$0.02 & 0.80$\pm$0.04 & 0.79$\pm$0.03 & 0.83$\pm$0.03 & 0.024$\pm$0.008 & 3.97$\pm$0.05 & 1.48$\pm$0.00 & 10.83$\pm$0.33 \\
\quad neural decoder & 0.35$\pm$0.01 & 0.18$\pm$0.01 & 0.75$\pm$0.03 & 0.81$\pm$0.04 & 0.138$\pm$0.005 & 3.95$\pm$0.07 & 1.48$\pm$0.01 & 10.35$\pm$0.17 \\
\quad GRU transition & 0.79$\pm$0.03 & 0.84$\pm$0.03 & 0.78$\pm$0.03 & 0.82$\pm$0.03 & 0.006$\pm$0.001 & 3.82$\pm$0.00 & 1.48$\pm$0.01 & 10.56$\pm$0.33 \\
\quad no action & 0.42 & 0.13 & 0.52 & 0.10 & 0.537 & 6.44 & 1.76 & 6.44 \\
\quad no rollout loss & 0.77 & 0.75 & 0.74 & 0.72 & 0.178 & 7.19 & 2.07 & 12.13 \\
\quad black box & -0.05$\pm$0.13 & -0.10$\pm$0.15 & 0.76$\pm$0.06 & 0.81$\pm$0.05 & -0.025$\pm$0.040 & 3.86$\pm$0.00 & 1.48$\pm$0.01 & 10.59$\pm$0.32 \\
\rowcolor{tabbase} History Transformer & -- & -- & 0.67 & 0.72 & -- & 4.16 & 1.47 & 10.53 \\
\midrule
\rowcolor{tabbase} Rain-history ridge (no model) & -- & -- & 0.73$\pm$0.03 & 0.75$\pm$0.03 & -- & -- & -- & -- \\
\bottomrule
\end{tabular}
}
\end{table*}

\begin{table*}[t]
\centering
\caption{\textbf{Twin experiment: true counterfactuals.}
All rain is removed after the last assimilation step; truth is re-simulated.
Rain effect: correlation and regression slope of the model's $\theta(\text{actual})-\theta(\text{no rain})$ on the true effect.
Drydown $\tau$: per-cell e-folding time of the zero-rain rollout (median) and its rank correlation $\rho$ with the truth.
Direct: designated surface storage; probe: ridge probe.}
\label{tab:twin_cf}
\resizebox{\textwidth}{!}{% generated by code/scripts/make_twin_table.py -- do not edit by hand
\begin{tabular}{l cccc cccc}
\toprule
\rowcolor{tabhead}  & \multicolumn{4}{c}{Direct readout} & \multicolumn{4}{c}{Probe readout} \\
\cmidrule(lr){2-5}\cmidrule(lr){6-9}
\rowcolor{tabhead}  & \multicolumn{2}{c}{Rain effect} & \multicolumn{2}{c}{Drydown $\tau$} & \multicolumn{2}{c}{Rain effect} & \multicolumn{2}{c}{Drydown $\tau$} \\
\cmidrule(lr){2-3}\cmidrule(lr){4-5}\cmidrule(lr){6-7}\cmidrule(lr){8-9}
\rowcolor{tabhead} Model & $R$ & slope & median (d) & $\rho$ & $R$ & slope & median (d) & $\rho$ \\
\midrule
\rowcolor{tabours} \method & 0.99$\pm$0.00 & 1.08$\pm$0.03 & 3.0$\pm$0.1 & 0.97$\pm$0.01 & 0.99$\pm$0.00 & 0.97$\pm$0.01 & 2.7$\pm$0.2 & 0.98$\pm$0.00 \\
\quad no residual content & 0.98$\pm$0.01 & 1.15$\pm$0.05 & 4.8$\pm$0.1 & 0.75$\pm$0.07 & 0.98$\pm$0.01 & 0.97$\pm$0.01 & 3.7$\pm$0.3 & 0.78$\pm$0.09 \\
\quad neural decoder & 0.04$\pm$0.08 & 0.02$\pm$0.03 & 3.6$\pm$4.1 & 0.49$\pm$0.35 & 0.97$\pm$0.01 & 0.95$\pm$0.02 & 2.9$\pm$0.4 & 0.90$\pm$0.03 \\
\quad GRU transition & 0.99$\pm$0.00 & 1.31$\pm$0.11 & 2.6$\pm$0.3 & 0.99$\pm$0.00 & 1.00$\pm$0.00 & 0.99$\pm$0.00 & 2.7$\pm$0.3 & 0.98$\pm$0.00 \\
\quad no action & -0.00 & -0.00 & 2.6 & -0.10 & -0.00 & -0.00 & 2.0 & -0.11 \\
\quad no rollout loss & 0.93 & 1.34 & 2.6 & 0.91 & 0.94 & 0.97 & 5.8 & 0.62 \\
\quad black box & -0.19$\pm$0.44 & -0.03$\pm$0.09 & 4.2$\pm$1.2 & 0.33$\pm$0.08 & 0.99$\pm$0.00 & 1.01$\pm$0.01 & 2.6$\pm$0.4 & 0.97$\pm$0.00 \\
\rowcolor{tabbase} History Transformer & -- & -- & -- & -- & 0.89 & 0.80 & 2.7 & 0.95 \\
\midrule
\rowcolor{tabbase} Rain-history ridge (no model) & -- & -- & -- & -- & 0.79$\pm$0.04 & 0.65$\pm$0.07 & 2.8$\pm$0.4 & 0.40$\pm$0.02 \\
\midrule
\rowcolor{tabbase} Truth & 1 & 1.00 & 2.7$\pm$0.2 & 1 & 1 & 1.00 & 2.7$\pm$0.2 & 1 \\
\bottomrule
\end{tabular}
}
\end{table*}

\section{Additional experiments}
\label{sec:supp_extra}
\paragraph{Latent precipitation.}
\label{sec:latentrain}
Satellite precipitation has multiplicative errors and misses storms, so a model that trusts it must explain every rain error as a state error.
As a variant, we treat the true action as latent: after observing $\obs_t$, the assimilator also corrects the precipitation of the previous step,
\begin{equation}
\tilde a_{t-1}=a_{t-1}\,e^{\delta_{t-1}}+\rho_{t-1},\qquad |\delta|\le2,\ \rho\ge0 ,
\end{equation}
and priors on rain and state errors separate the two, as in weak-constraint variational assimilation~\cite{tremolet2006weak}.
The real-data runs use the given precipitation; this variant is tested only in the twin.
The latent-precipitation experiment corrupts the given rain with log-normal multiplicative errors ($\sigma=0.7$) and missed storms (about 13\% of rainy cells) and compares the corrected rain with the truth.
A control run on clean rain checks that the correction does not invent rain.
Latent precipitation makes the state robust to rain errors but does not retrieve the rain (\cref{tab:h3}).
With corrupted rain, it lowers the soil-moisture bias from $+0.040$ to $+0.016$--$0.024$ and the ubRMSE from 0.0166 to 0.0137--0.0149 under either prior, while the corrected rain is no closer to the truth (total bias $-17$ to $-19\%$ against $-13\%$).
On clean rain it adds only 2--3\%.
The observations constrain the consequences of a rain error more tightly than the rain itself, so the correction acts as a learned forcing-error term rather than a rain retrieval.
In the latent-precipitation model, the transition $t-1\to t$ is re-run from the previous analysis with $\tilde a_{t-1}$, and the state increment is applied on top.
The data alone cannot separate a rain error from a state error; as in weak-constraint variational assimilation~\cite{tremolet2006weak}, priors on both errors resolve this:
\begin{itemize}
\item a log-normal rain error, $\delta\sim\mathcal{N}(0,\sigma_a^2)$ with $\sigma_a=0.7$;
\item an exponential prior on added (missed) rain, with a mean of 2\,mm per step;
\item a Gaussian prior on the per-step log change of the physical storages, $\sigma_S=0.1$.
\end{itemize}
The negative log-priors are summed over cells and divided by the mean number of observation tokens per sensor.
Each prior element thus weighs as much as one average observation, and the priors are not counted as an extra sensor.
Observations are still scored under the uncorrected prior state, so the correction cannot leak into the score of the observations that produced it.
The corrections are zero-initialized, so the model starts exactly as without latent rain.
We also report a tighter setting, with an added-rain mean of 0.5\,mm and $\sigma_S=0.2$.
In the twin experiment the model receives a corrupted precipitation field, and we test both the state and whether $\tilde a$ moves towards the true rain.

\Cref{tab:h3} gives the twin results for both prior settings.
The given rain is corrupted with log-normal multiplicative errors ($\sigma=0.7$) and missed storms; the clean rows are the control.
For runs with latent precipitation, the rain columns score the corrected rain $\tilde a$.

\begin{table}[h]
\centering
\caption{\textbf{Latent precipitation in the twin.} Filtered surface soil moisture against the truth, and the rain used by the model against the true rain (8 test windows).}
\label{tab:h3}
\resizebox{\linewidth}{!}{% generated by code/scripts/make_h3_table.py -- do not edit by hand
\begin{tabular}{lll ccc cc}
\toprule
\rowcolor{tabhead}  & & & \multicolumn{3}{c}{Filtered $\theta_s$} & \multicolumn{2}{c}{Rain used (vs.\ truth)} \\
\cmidrule(lr){4-6}\cmidrule(lr){7-8}
\rowcolor{tabhead} Rain given & Latent & Prior & $R$ & ubRMSE & bias & daily RMSE (mm) & total bias \\
\midrule
corrupted & -- & -- & 0.597 & 0.0166 & +0.040 & 2.78 & -12.6\% \\
corrupted & yes & default & 0.613 & 0.0149 & +0.016 & 2.82 & -18.5\% \\
corrupted & yes & tight & 0.627 & 0.0137 & +0.024 & 2.94 & -17.3\% \\
\midrule
clean & yes & default & 0.711 & 0.0071 & +0.006 & 0.24 & +2.1\% \\
clean & yes & tight & 0.724 & 0.0069 & +0.006 & 0.25 & +2.7\% \\
\bottomrule
\end{tabular}
}
\end{table}

\paragraph{Rain-pulse response.}
We add synthetic rain pulses to the forcing of a filtered state and compare the soil-moisture increment with natural events of similar size and antecedent state.
Natural events are isolated 6\,h gauge totals at the stations (less than 0.5\,mm in the 24\,h before and after); their increment is the change of the 6\,h station mean across the rain step.
Synthetic pulses of 2.5, 7, 15 and 30\,mm are added at dry times in the station cells, and their increment is the difference between rollouts with and without the pulse over the same 6\,h.
Both are grouped by rain amount (1--4, 4--10, 10--20 and 20--50\,mm) and by the tercile of the station's antecedent value.
On the validation year, over 300 natural events and 7784 pulses, the group medians of \method and the stations are rank-correlated with $\rho=0.86$ in each of three seeds (11 groups with at least 10 natural events).
The variant without rollout loss gives 0.86 and 0.92 but responds too strongly (\cref{tab:real_sm}), and without the action the model does not respond to rain at all (two seeds).
The GRU transition gives $\rho=0.54$, 0.43 and 0.20 in its three seeds, and its increments are at most 0.003 in every group, against 0.05--0.14 for the stations from 10\,mm on and up to 0.13--0.15 for \method: although its forecasts depend on the rain nearly as much as those of \method (zeroing the rain raises the SMAP NLL by $2.98\pm0.06$, against $3.31\pm0.10$), it barely moves the storage in the step in which an isolated pulse falls (for rain over the whole domain, see \cref{sec:local}).
For seed 0, from 10\,mm on the increments agree in size (0.062 against 0.062 for 10--20\,mm, 0.117 against 0.108 for 20--50\,mm).
Within every rain group the model's increment decreases with antecedent wetness, as wetter soil can store less.

\paragraph{Held-out sensor.}
During training, each sensor is withheld from assimilation with probability 0.2 (\cref{sec:method}).
We test whether this yields a state that does not depend on a particular sensor: on the validation year, the three seeds of \method filter with one sensor never assimilated, while everything else, including the evaluation, is unchanged (\cref{tab:heldout}).
Without CYGNSS the scores are essentially unchanged.
Without SMAP, CYGNSS specular points alone give a station correlation of 0.80 and an anomaly correlation of 0.67, against 0.82 and 0.71 with both sensors and 0.79 and 0.68 for SMAP Level-4, which assimilates SMAP brightness temperatures.
Paired over stations (Wilcoxon signed-rank test, each seed separately), CYGNSS-only filtering is indistinguishable from SMAP Level-4 in both scores (better at 14--16 of 28 stations, $p\ge0.34$) and better than ERA5-Land ($p\le0.016$) and the LSTM on SMAP retrievals in correlation ($p\le0.001$).
Assimilating SMAP alone gives the same scores as both sensors, because SMAP leaves no gaps for CYGNSS to fill here: it observes every station cell at least every two days (median 18\,h since the last overpass, against 72\,h for CYGNSS); in the 23\% of steps one to two days after an overpass, adding CYGNSS raises the anomaly correlation only slightly (0.738 against 0.725, seed mean).

\section{Twin with a structural mismatch}
\label{sec:supp_mm}
The truth of the main twin differs from \method in functional form but, like it, moves water only vertically.
To test structure \method cannot represent, a second twin adds three processes to the same worlds (same parameters, forcing and observation operators): the excess that does not pond flows to the lowest of the four neighbours and arrives one step later; root-zone water diffuses laterally between neighbours (5\% of the difference per step); and 20\% of the cells, in a smooth mask, are irrigated with 4\,mm per step whenever the surface is below its critical moisture under high evaporative demand.
Irrigation is not part of the action, so the models must explain it from the observations, and it also continues in the rain-free counterfactual.
We train the same configurations as in the main twin (\method with two seeds, one seed otherwise) and evaluate them identically (\cref{tab:twin_mm}).
All models forecast the observations about equally well (rollout SMAP NLL 4.17--4.32; History Transformer 4.56), and a probe recovers soil moisture (0.76--0.77) and the effect of rain ($R\ge0.96$) from all four (History Transformer: 0.69 and 0.86).
Read directly, only the physical decoders expose both: state $R$ 0.78 and 0.80 against 0.20 and $-0.26$, effect slope 0.98 and 1.21 against 0.29 and $-0.13$.
\method loses little against the main twin (state $R$ 0.78 against 0.83, effect $R$ 0.97 against 0.99, median drydown $\tau$ 3.2 against a true 2.9\,days).

\begin{table*}[t]
\centering
\caption{\textbf{Twin with a structural mismatch} (lateral run-on, lateral subsurface flow, unobserved irrigation). Columns as in \cref{tab:twin}.}
\label{tab:twin_mm}
\resizebox{\textwidth}{!}{% generated by code/scripts/make_twin_table.py -- do not edit by hand
\begin{tabular}{l cc c ccc ccc}
\toprule
\rowcolor{tabhead}  & \multicolumn{2}{c}{State $R$ (rollout)} & Rollout & \multicolumn{3}{c}{Rain effect, direct} & \multicolumn{3}{c}{Rain effect, probe} \\
\cmidrule(lr){2-3}\cmidrule(lr){5-7}\cmidrule(lr){8-10}
\rowcolor{tabhead} Model & direct & probe & SMAP NLL & $R$ & slope & $\tau$ $\rho$ & $R$ & slope & $\tau$ $\rho$ \\
\midrule
\rowcolor{tabours} \method & 0.78$\pm$0.04 & 0.76$\pm$0.03 & 4.26$\pm$0.03 & 0.97$\pm$0.00 & 0.98$\pm$0.06 & 0.93$\pm$0.01 & 0.97$\pm$0.00 & 0.89$\pm$0.00 & 0.90$\pm$0.00 \\
\quad neural decoder & 0.20 & 0.77 & 4.32 & 0.45 & 0.29 & 0.86 & 0.96 & 0.88 & 0.86 \\
\quad GRU transition & 0.80 & 0.77 & 4.17 & 0.97 & 1.21 & 0.98 & 0.98 & 0.95 & 0.98 \\
\quad black box & -0.26 & 0.76 & 4.18 & -0.23 & -0.13 & 0.32 & 0.97 & 0.95 & 0.96 \\
\rowcolor{tabbase} History Transformer & -- & 0.69 & 4.56 & -- & -- & -- & 0.86 & 0.75 & 0.91 \\
\midrule
\rowcolor{tabbase} Rain-history ridge (no model) & -- & 0.70$\pm$0.04 & -- & -- & -- & -- & 0.77$\pm$0.04 & 0.54$\pm$0.02 & 0.36$\pm$0.11 \\
\bottomrule
\end{tabular}
}
\end{table*}

\section{Twin with mis-specified operators}
\label{sec:supp_ms}
In both twins above, observations are generated with the forward models that the physical decoders use, so a grounded decoder only has to invert a correctly specified operator.
A third twin keeps the bucket truth of \cref{sec:supp_twin} and generates the observations with operators outside the decoders' family (\texttt{synthetic.py}, \texttt{obs="misspec"}).
For SMAP: the dielectric model of Dobson~\etal instead of Mironov~\etal; polarization mixing ($Q=0.15$); a roughness that decreases with soil moisture and varies in space; angular exponents of 1 for both polarizations; a vegetation optical depth with a spatially varying $b$ parameter on vegetation lagged by 30 days; a seasonal single-scattering albedo; and an effective temperature that mixes surface and daily-mean temperature according to soil moisture.
For CYGNSS: the same dielectric model, an incoherent component (5\% of the roughness-scattered power) and vegetation attenuation scaled to L1.
Relative to the decoders' operators on the same states, this changes H- and V-polarized brightness temperatures by 6.4 and 13.6\,K (rms; $-12.8$\,K mean for V) and reflectivity by 2.1\,dB (rms), against noise of 1.5\,K and 1\,dB; the noise draws, sampling and truth are identical to the matched twin.
We train the same configurations as in the structural-mismatch twin and evaluate them identically (\cref{tab:twin_ms}).
The observation likelihood barely tells the models apart: rollout SMAP NLL is 3.96 for \method, 3.83--3.90 for the other models and 4.03 for the History Transformer, so the mis-specified physics costs \method only a little likelihood.
A probe again recovers soil moisture (0.77--0.79, History Transformer 0.71) and the effect of rain ($R\ge0.98$) from every model.
Read directly, only the physical decoders expose them: state $R$ $0.81\pm0.03$ for \method and 0.79 with a GRU transition, against 0.00 with a neural decoder and $-0.18$ for the black box, and effect $R$ 0.98 and 0.99 against 0.05 and $-0.03$.
The price of the wrong operators lies in magnitudes rather than in readability: \method overestimates the effect more than in the matched twin (slope $1.25\pm0.02$ against $1.08\pm0.03$) and dries more slowly, while still ranking the drydowns ($\rho=0.87\pm0.02$).
Grounding thus does not require exact operators; a decoder of the right family, with bounded learned parameters, still pins the meaning of the state.

\begin{table*}[t]
\centering
\caption{\textbf{Twin with mis-specified observation operators} (outside the decoders' family). \method: two seeds; columns as in \cref{tab:twin}.}
\label{tab:twin_ms}
\resizebox{\textwidth}{!}{% generated by code/scripts/make_twin_table.py -- do not edit by hand
\begin{tabular}{l cc c ccc ccc}
\toprule
\rowcolor{tabhead}  & \multicolumn{2}{c}{State $R$ (rollout)} & Rollout & \multicolumn{3}{c}{Rain effect, direct} & \multicolumn{3}{c}{Rain effect, probe} \\
\cmidrule(lr){2-3}\cmidrule(lr){5-7}\cmidrule(lr){8-10}
\rowcolor{tabhead} Model & direct & probe & SMAP NLL & $R$ & slope & $\tau$ $\rho$ & $R$ & slope & $\tau$ $\rho$ \\
\midrule
\rowcolor{tabours} \method & 0.81$\pm$0.03 & 0.81$\pm$0.03 & 3.96$\pm$0.00 & 0.98$\pm$0.00 & 1.25$\pm$0.02 & 0.87$\pm$0.02 & 0.98$\pm$0.00 & 0.94$\pm$0.01 & 0.86$\pm$0.04 \\
\quad neural decoder & 0.00 & 0.77 & 3.90 & 0.05 & 0.03 & 0.49 & 0.98 & 0.96 & 0.87 \\
\quad GRU transition & 0.79 & 0.79 & 3.83 & 0.99 & 1.33 & 0.98 & 0.99 & 0.97 & 0.97 \\
\quad black box & -0.18 & 0.79 & 3.86 & -0.03 & -0.01 & 0.27 & 0.99 & 0.97 & 0.92 \\
\rowcolor{tabbase} History Transformer & -- & 0.71 & 4.03 & -- & -- & -- & 0.89 & 0.78 & 0.96 \\
\midrule
\rowcolor{tabbase} Rain-history ridge (no model) & -- & 0.73$\pm$0.02 & -- & -- & -- & -- & 0.81$\pm$0.03 & 0.61$\pm$0.03 & 0.39$\pm$0.03 \\
\bottomrule
\end{tabular}
}
\end{table*}

\section{Probing the real-data latent}
\label{sec:supp_probe}
The direct readout asks whether the designated surface storage \emph{is} soil moisture, without any fitting.
To ask whether soil moisture can be \emph{decoded} from the latent, as the twin probe does, we fit a ridge regression from the normalized prior latent $[\bm h,\bm u]$ (101 features) in the station's cell to the station value, on filtering states of the training period (August 2018--December 2023); the regularization is chosen on 2023 after fitting on the earlier years, and one probe serves all stations.
It is then applied, unchanged, on the same validation and test windows as the direct readout, while filtering and along the 10-day rollouts (\cref{tab:probe_real}).
The probe reads soil moisture from the neural-decoder and black-box latents as well as from \method, whose storage reaches the same correlation without any fit.
Along rollouts, the probe trained on filtering states transfers less well to the neural-decoder model (10-day $R=0.59\pm0.27$ on the validation year), whose latent drifts from the states it was fitted on.

\begin{table*}[t]
\centering
\caption{\textbf{Direct readout versus a linear probe on real data.} Direct: $\theta_s=S_s/50$\,mm in the station's cell, no fitting (for the neural-decoder and black-box models, the designated channel). Probe: ridge regression from the latent to the station value, fitted on the training period. Median per-station correlation; mean $\pm$ std over seeds.}
\label{tab:probe_real}
\resizebox{\textwidth}{!}{% generated by code/scripts/make_probe_table.py -- do not edit by hand
\begin{tabular}{l c cc cc cc cc}
\toprule
\rowcolor{tabhead}  & & \multicolumn{4}{c}{Validation year} & \multicolumn{4}{c}{Test year} \\
\cmidrule(lr){3-6}\cmidrule(lr){7-10}
\rowcolor{tabhead}  & & \multicolumn{2}{c}{Filtering $R$} & \multicolumn{2}{c}{10-day forecast $R$} & \multicolumn{2}{c}{Filtering $R$} & \multicolumn{2}{c}{10-day forecast $R$} \\
\cmidrule(lr){3-4}\cmidrule(lr){5-6}\cmidrule(lr){7-8}\cmidrule(lr){9-10}
\rowcolor{tabhead} Model & seeds & direct & probe & direct & probe & direct & probe & direct & probe \\
\midrule
\rowcolor{tabours} \method & 3 & 0.82$\pm$0.00 & 0.82$\pm$0.01 & 0.80$\pm$0.01 & 0.77$\pm$0.04 & 0.71$\pm$0.01 & 0.71$\pm$0.01 & 0.66$\pm$0.01 & 0.65$\pm$0.02 \\
\quad neural decoder & 2 & 0.32$\pm$0.02 & 0.81$\pm$0.01 & 0.31$\pm$0.08 & 0.59$\pm$0.27 & 0.37$\pm$0.03 & 0.69$\pm$0.04 & 0.28$\pm$0.05 & 0.46$\pm$0.13 \\
\quad black box & 1 & -0.04 & 0.81 & 0.08 & 0.78 & -0.07 & 0.73 & -0.02 & 0.65 \\
\bottomrule
\end{tabular}
}
\end{table*}

\section{Rain-history baseline on real data}
\label{sec:supp_api}
The twin's model-free reference is a ridge regression from antecedent-precipitation features.
We fit its real-data counterpart to the probes: causal exponential filters of the IMERG rain of the station's cell with time constants of 6\,h to 30\,days and of MERRA-2 air temperature and shortwave radiation (1 and 10\,days), using forcing before the state time only.
The ridge is fitted on the training period (August 2018--December 2023, regularization chosen on 2023), either once for all stations or per station; \cref{tab:real_sm,tab:test_sm} report the stronger per-station fit.
Its series is scored on the samples at which \method is scored; since it depends on rain only, the same series serves as its 10-day forecast.
On the validation year, the per-station ridge reaches $R=0.79$ and anomaly $R=0.71$ (one ridge for all stations: 0.78 and 0.71), against 0.82 and 0.71 for \method, which is better at 18--19 of 28 stations; on the test year, 0.64 and 0.69 (0.57 and 0.71) against 0.71 and 0.76, better at 13--14 of 18 stations.
Fitted to the probes, a regression on rain history is a strong station predictor; but it can only follow the rain it is given, and where IMERG misses or invents a storm, \method, which never sees the probes, comes closer to them (\cref{sec:supp_mismatch}).

\section{Same-sensor retrievals, level and amplitude}
\label{sec:supp_retr}
With $\theta_s=S_s/50$\,mm and physical operators, the direct readout is a retrieval regularized by the learned dynamics.
Its no-dynamics counterpart inverts each overpass separately.
We score three such products at the stations, each on its own samples and against the same centred in-situ value as \method (\cref{tab:retrieval}).
SMAP Level-3 (enhanced, 9\,km, the dual-channel retrieval of the AM and PM passes) is assigned to the state time nearest to its brightness-temperature time; with the recommended quality flag, 8 of the 28 stations have no retrieval in the validation year, so we also report all retrievals.
CYGNSS Level-3 soil moisture (version 3.2, 9\,km, 6-hour slices) regresses angle-normalized reflectivity on SMAP soil moisture, calibrated on August 2018--November 2023; the value at a state time is the mean of the two adjacent slices, as for the probes.
The process-guided retrieval PGML~\cite{feng2026pgml} maps SMAP brightness temperatures, vegetation, soil and weather to soil moisture with a network pretrained on radiative-transfer simulations and fine-tuned to in-situ probes of April 2015--December 2023, among them ISMN, which includes the SCAN and USCRN networks used here; it gives one value per day at the 6 AM overpass, which we assign to 12 UTC, and flags every cell of the region with its dense-vegetation bit, so we keep all values.
\method is scored on the same samples, assimilating both sensors or only the retrieval's sensor.
On the validation year, \method has the higher correlation at most stations against both Level-3 products ($p\le0.006$ per seed, Wilcoxon, uncorrected), and with CYGNSS alone it improves on the CYGNSS retrieval by 0.2 in correlation at 25--26 of 26 stations; on the test year it is better than CYGNSS Level-3 at all 16 stations, than all SMAP Level-3 retrievals in median correlation (0.72 against 0.66), and than recommended SMAP Level-3 retrievals at 11 of 13 stations, with a smaller ubRMSE (0.032 against 0.038).
Against PGML, which has seen the probes of these networks in earlier years, \method is level without any label: $R=0.82$ against 0.81 in the validation year, 0.66--0.69 against 0.67 in the test year.

Correlation and ubRMSE ignore an affine map between model and probe, so the scale fixed by the operators (and the 50\,mm depth) is not tested by them.
Per station on the validation year, the median bias of \method is $-0.009$, $+0.003$ and $-0.016$ in the three seeds (interquartile range about $\pm0.04$), and the ratio of standard deviations model/probe 0.98, 1.16 and 1.01; on the same samples, SMAP Level-4 has $+0.002$ and 0.88 (test year: \method $-0.002$ to $+0.007$ and 0.90--1.08; SMAP Level-4 $+0.006$ and 1.02).
The readout thus has the probes' level and amplitude about as well as SMAP Level-4, without any fit.

\begin{table*}[t]
\centering
\caption{\textbf{\method against per-overpass retrievals of the same sensors}, scored on the retrievals' samples (median over stations; bias = model $-$ probe). \method: mean $\pm$ std over three seeds. Better: stations at which \method (both sensors / only the retrieval's sensor) has the higher correlation than the retrieval, range over seeds. $^\dagger$Fine-tuned to in-situ probes (ISMN, including SCAN and USCRN) of 2015--2023; one value per day.}
\label{tab:retrieval}
\resizebox{\textwidth}{!}{% generated by code/scripts/make_retrieval_table.py -- do not edit by hand
\begin{tabular}{l cc ccc ccc ccc c}
\toprule
\rowcolor{tabhead}  & & & \multicolumn{3}{c}{Retrieval} & \multicolumn{3}{c}{\method, both sensors} & \multicolumn{3}{c}{\method, same sensor only} & Better \\
\cmidrule(lr){4-6}\cmidrule(lr){7-9}\cmidrule(lr){10-12}
\rowcolor{tabhead} Retrieval & Stations & Samples & $R$ & ubRMSE & bias & $R$ & ubRMSE & bias & $R$ & ubRMSE & bias & both / same \\
\midrule
\multicolumn{13}{l}{\emph{Validation year}} \\
SMAP L3 (recommended) & 20 & 4745 & 0.72 & 0.044 & -0.040 & 0.77$\pm$0.01 & 0.039$\pm$0.003 & -0.021$\pm$0.005 & 0.77$\pm$0.00 & 0.040$\pm$0.002 & -0.020$\pm$0.004 & 17 / 17 \\
SMAP L3 (all) & 28 & 7703 & 0.76 & 0.050 & -0.015 & 0.82$\pm$0.00 & 0.045$\pm$0.001 & -0.008$\pm$0.009 & 0.81$\pm$0.00 & 0.045$\pm$0.001 & -0.009$\pm$0.009 & 21--22 / 21--22 \\
CYGNSS L3 & 26 & 5743 & 0.61 & 0.055 & -0.014 & 0.84$\pm$0.01 & 0.044$\pm$0.002 & -0.001$\pm$0.007 & 0.82$\pm$0.01 & 0.045$\pm$0.002 & -0.010$\pm$0.003 & 26 / 25--26 \\
SMAP PGML$^\dagger$ & 28 & 3860 & 0.81 & 0.042 & -0.008 & 0.82$\pm$0.00 & 0.045$\pm$0.001 & -0.007$\pm$0.008 & 0.82$\pm$0.00 & 0.045$\pm$0.001 & -0.007$\pm$0.008 & 15--16 / 14 \\
\midrule
\multicolumn{13}{l}{\emph{Test year}} \\
SMAP L3 (recommended) & 13 & 2540 & 0.54 & 0.038 & -0.018 & 0.50$\pm$0.02 & 0.032$\pm$0.003 & -0.012$\pm$0.001 & 0.50$\pm$0.01 & 0.031$\pm$0.003 & -0.011$\pm$0.001 & 11 / 11 \\
SMAP L3 (all) & 18 & 4033 & 0.66 & 0.048 & -0.002 & 0.72$\pm$0.01 & 0.038$\pm$0.002 & 0.001$\pm$0.004 & 0.72$\pm$0.01 & 0.038$\pm$0.002 & 0.001$\pm$0.005 & 12 / 12 \\
CYGNSS L3 & 16 & 2871 & 0.44 & 0.042 & +0.001 & 0.64$\pm$0.00 & 0.037$\pm$0.003 & 0.007$\pm$0.004 & 0.63$\pm$0.00 & 0.036$\pm$0.002 & 0.004$\pm$0.004 & 16 / 16 \\
SMAP PGML$^\dagger$ & 18 & 1294 & 0.67 & 0.037 & +0.003 & 0.67$\pm$0.02 & 0.035$\pm$0.004 & 0.014$\pm$0.005 & 0.67$\pm$0.02 & 0.035$\pm$0.003 & 0.012$\pm$0.006 & 9 / 9 \\
\bottomrule
\end{tabular}
}
\end{table*}

\section{One state, not one of many}
\label{sec:supp_seeds}
Physical operators give the state coordinates, but with 96 free channels and learned operator parameters, could several combinations of storage and parameters explain the same observations?
If so, independently trained models would settle on different ones.
We compare the filtered surface soil moisture of independently trained seeds over the whole domain on the validation year, not only at the stations: per 9\,km cell, the temporal correlation, the mean difference and the ratio of standard deviations, summarised over the 10,260 cells (\texttt{scripts/diag/seed\_consistency.py}).
The three seeds of \method agree almost exactly: median per-cell correlation 0.992--0.993 (interquartile range 0.988--0.995), with more than 99.9\% of cells above 0.9, and their means differ by a median 0.006--0.011.
With a GRU transition, the three seeds agree nearly as closely (0.985--0.992), and the first agrees with \method (0.96).
With neural decoders, the designated channel agrees less between seeds (0.87; 27\% of cells above 0.9) and hardly with \method (0.35): without the operators, each seed finds its own solution, which is why only a probe fitted to the stations can read them.
The physical operators thus pin the state to one solution.

\section{IMERG--gauge disagreements}
\label{sec:supp_mismatch}
The rain-history ridge (\cref{sec:supp_api}) matches \method at the stations, so station correlation cannot separate a model of the soil from a filter of the rain.
They must part where the rain they are given is wrong.
IMERG and the stations' gauges agree on most days (6-hour correlation 0.70 at USCRN and 0.65 at SCAN stations, highest at lag zero), but not on all.
We select station-days of the validation and test years on which the gauge records at least a threshold and IMERG in the station's cell less than 2\,mm (IMERG missed), or IMERG at least the threshold and the gauge less than 1\,mm (IMERG false alarm); SCAN gauges report no data in 2025.
For each, we compare the change of soil moisture from the state time before the day to the mean of the following day: of the probe, of \method while filtering, and of the ridge (\texttt{scripts/diag/rain\_mismatch.py}).
The ridge sees only IMERG and follows it: after missed storms of at least 10\,mm it does not wet the soil, and after false alarms it does (\cref{tab:mismatch}).
\method is driven by the same IMERG rain, but its filtered state moves toward the probe: its change is closer to the probe's than the ridge's at 26--28 of 36 missed storms and at 31--36 of 54 false alarms, and at 9--10 of 11 and 14--15 of 19 with a 20\,mm threshold.
On days when IMERG and the gauge agree, \method and the ridge are equally close to the probe: the separation appears where the rain is wrong and the satellites can correct it.

\section{Decoder bounds and inputs}
\label{sec:supp_sens}
The physical decoders carry two kinds of bounds: the neural residual added to their outputs ($\pm2$\,K and $\pm1$\,dB, a scale of 0.1) and the ranges of their learned physical parameters (the learned vegetation-opacity term, single-scattering albedo up to 0.12, roughness, an effective-temperature correction of at most $\pm5$\,K, and CYGNSS surface roughness of 0.5--5.5\,cm).
To test how much the real-data results rest on these choices, we retrain seed 0 of \method with the residual removed, with its bound tripled ($\pm6$\,K and $\pm3$\,dB), or with all parameter ranges doubled, keeping everything else fixed, and evaluate as in \cref{tab:real_sm,tab:real_obs} (\cref{tab:sens_real}).
Removing or tripling the residual changes filtering and forecast correlations by at most 0.01, the rollout NLL by at most 0.02 and the zero-rain $\Delta$NLL (3.44 and 3.36) within the spread of the three default seeds; the response to synthetic rain pulses keeps its ranking over cells (rank correlation 0.86 in every run) and its size.
Doubling the parameter ranges leaves the likelihoods unchanged but weakens the state: the 10-day correlation drops to 0.76 and ubRMSE rises to 0.048.
The residual thus does not carry the signal, and the results do not hinge on its bound; the physical ranges matter more, since wider ones let the decoder explain the same observations with a less constrained storage.

The decoders also read the free channels, which set roughness and vegetation opacity.
Could these channels, rather than the storage, explain the signals, so that the storage is a readout only in name?
We retrain two seeds whose decoders see the storage, the static inputs and the viewing geometry but not the free channels (\texttt{--no\_dec\_free}).
The storage then has to explain the observations alone, and it remains soil moisture without any fit to the probes: station correlation $0.78\pm0.01$ (test year 0.63--0.65), against 0.82 with free channels and 0.32 with neural decoders.
What the model loses is mostly fit: the rollout NLL rises from 5.41 to 5.98 for SMAP, and the zero-rain $\Delta$NLL falls to 2.46 while remaining large.
Its storage also overreacts to rain: synthetic 10--20\,mm pulses raise it by a median 0.11 (default seeds 0.06--0.07; natural storms of that size 0.06 at the probes), with the ranking over cells unchanged (0.86).
The free channels do not stand in for the storage; they absorb the vegetation and roughness changes that the storage would otherwise have to carry.

\begin{table*}[t]
\centering
\caption{\textbf{Sensitivity to the decoder bounds and inputs} (real data, validation year). Seed 0 of \method retrained with one bound changed, and decoders that do not read the free channels (mean $\pm$ std over two seeds); the last row gives mean $\pm$ std over the three seeds of the default model. Columns as in \cref{tab:real_sm,tab:real_obs}.}
\label{tab:sens_real}
\resizebox{\textwidth}{!}{% generated by code/scripts/make_sens_table.py -- do not edit by hand
\begin{tabular}{l ccc c cc cc c}
\toprule
\rowcolor{tabhead}  & \multicolumn{3}{c}{Filtering} & 10-day & \multicolumn{2}{c}{Response to rain} & \multicolumn{2}{c}{Rollout NLL} & Zero-rain \\
\cmidrule(lr){2-4}\cmidrule(lr){6-7}\cmidrule(lr){8-9}
\rowcolor{tabhead} Decoders & $R$ & ubRMSE & anomaly $R$ & $R$ & 10--20\,mm & 20--50\,mm & SMAP & CYGNSS & SMAP $\Delta$NLL \\
\midrule
\rowcolor{tabours} \method (residual 0.1, ranges $\times1$; seed 0) & 0.82 & 0.045 & 0.70 & 0.80 & 0.062 & 0.117 & 5.40 & 2.05 & 3.35 \\
\quad no residual & 0.81 & 0.044 & 0.70 & 0.79 & 0.079 & 0.149 & 5.42 & 2.04 & 3.44 \\
\quad residual 0.3 & 0.81 & 0.045 & 0.70 & 0.79 & 0.070 & 0.129 & 5.41 & 2.05 & 3.36 \\
\quad parameter ranges $\times2$ & 0.80 & 0.048 & 0.71 & 0.76 & 0.064 & 0.121 & 5.40 & 2.05 & 3.28 \\
\quad decoders without free channels (two seeds) & 0.78$\pm$0.01 & 0.051$\pm$0.001 & 0.67$\pm$0.01 & 0.78$\pm$0.01 & 0.113$\pm$0.003 & 0.169$\pm$0.001 & 5.98$\pm$0.02 & 2.12$\pm$0.00 & 2.46$\pm$0.13 \\
\midrule
\rowcolor{tabours} \method, three seeds & 0.82$\pm$0.00 & 0.045$\pm$0.001 & 0.71$\pm$0.01 & 0.80$\pm$0.01 & 0.069$\pm$0.005 & 0.128$\pm$0.010 & 5.41$\pm$0.01 & 2.05$\pm$0.01 & 3.31$\pm$0.10 \\
\bottomrule
\end{tabular}
}
\end{table*}

\section{Real-data drydowns and the root zone}
\label{sec:supp_dry}
\paragraph{Drydowns.}
For each rain-free drydown of the validation year at the stations, the model filters to the end of the rain and is then rolled out with all later rain removed, and e-folding times are fitted to model and probe in the same way.
On 274 drydowns, the model dries at the probes' rate on average (median ratio of e-folding times $0.96\pm0.14$ over seeds), but it does not rank events or stations (rank correlation 0.00--0.01 over events and $-0.33$ to $-0.40$ over stations): it dries fastest in the semi-arid west, where the probes dry slowest.
There the drying amplitude is close to probe noise and the point-to-cell mismatch is largest; event-scale drydowns are therefore verified only in the twin.

\paragraph{Root zone.}
Only the surface storage is constrained by L-band and GNSS-R observations, which sense the top few centimetres.
The root-zone storage, $S_r/950$\,mm, correlates with depth-weighted profile probes (10--100\,cm) at $R=0.48$ (seed 0, 26 stations), against 0.75 for SMAP Level-4, and its level is far too low: it acts as a reservoir for the surface dynamics, not as root-zone soil moisture.

\section{Test year}
\label{sec:supp_test}
All models, settings and the checkpoint rule were frozen before the test year (January--September 2025) was evaluated, once, with the same code as the validation year (\cref{tab:test_sm,tab:test_obs,tab:test_heldout}).
Only 18 of the 28 stations report in this period, and every product correlates less with them than on the validation year, including SMAP Level-4 (0.63 against 0.79) and ERA5-Land (0.56 against 0.71).
Paired over stations, \method has a higher anomaly correlation than SMAP Level-4 at 13--14 stations ($p\le0.003$ in each seed) and a higher correlation at 11--12 ($p\ge0.23$), and is better than ERA5-Land and the LSTM on SMAP retrievals in both scores ($p\le0.034$).
Filtering with CYGNSS alone is indistinguishable from SMAP Level-4 ($p\ge0.15$ in both scores and every seed).
All $p$-values are two-sided, per seed and not corrected for multiple comparisons.
In the rain-pulse test (6 groups with at least 10 natural events), the group medians of \method and the stations are rank-correlated with $\rho=0.77$--0.89, and the GRU transition again barely responds (\cref{tab:test_sm}).

\begin{table*}[t]
\centering
\caption{\textbf{Station soil moisture on the test year} (January--September 2025, median over 18 stations). Columns and protocol as in \cref{tab:real_sm}.}
\label{tab:test_sm}
\resizebox{\textwidth}{!}{% generated by code/scripts/make_real_table.py -- do not edit by hand
\begin{tabular}{l ccc ccc cc}
\toprule
\rowcolor{tabhead}  & \multicolumn{3}{c}{Filtering} & \multicolumn{3}{c}{Forecast $R$, lead (days)} & \multicolumn{2}{c}{Response to rain} \\
\cmidrule(lr){2-4}\cmidrule(lr){5-7}\cmidrule(lr){8-9}
\rowcolor{tabhead} Model & $R$ & ubRMSE & anomaly $R$ & 1 & 5 & 10 & 10--20\,mm & 20--50\,mm \\
\midrule
\rowcolor{tabours} \method & 0.71$\pm$0.01 & 0.038$\pm$0.002 & 0.76$\pm$0.01 & 0.73$\pm$0.01 & 0.65$\pm$0.01 & 0.66$\pm$0.01 & 0.071$\pm$0.005 & 0.133$\pm$0.007 \\
\quad GRU transition & 0.68$\pm$0.00 & 0.051$\pm$0.007 & 0.72$\pm$0.01 & 0.71$\pm$0.01 & 0.61$\pm$0.02 & 0.64$\pm$0.01 & 0.001$\pm$0.000 & 0.002$\pm$0.000 \\
\quad neural decoder & 0.37$\pm$0.03 & 0.053$\pm$0.007 & 0.49$\pm$0.05 & 0.29$\pm$0.02 & 0.41$\pm$0.01 & 0.28$\pm$0.05 & -- & -- \\
\quad no action & 0.64$\pm$0.01 & 0.034$\pm$0.000 & 0.62$\pm$0.03 & 0.66$\pm$0.00 & 0.48$\pm$0.02 & 0.41$\pm$0.01 & -- & -- \\
\quad no rollout loss & 0.70$\pm$0.00 & 0.043$\pm$0.001 & 0.76$\pm$0.00 & 0.73$\pm$0.00 & 0.58$\pm$0.04 & 0.59$\pm$0.05 & 0.093$\pm$0.002 & 0.161$\pm$0.010 \\
\midrule
\rowcolor{tabbase} Persistence (oracle) & -- & -- & -- & 0.94 & 0.58 & 0.24 & -- & -- \\
\rowcolor{tabbase} Probes, natural storms & -- & -- & -- & -- & -- & -- & 0.048 & 0.098 \\
\rowcolor{tabbase} Rain-history ridge (station-fit) & 0.64 & 0.033 & 0.69 & 0.63 & 0.59 & 0.64 & -- & -- \\
\rowcolor{tabbase} LSTM on SMAP L3 & 0.64 & 0.046 & 0.69 & -- & -- & -- & -- & -- \\
\rowcolor{tabbase} SMAP Level-4 & 0.63 & 0.043 & 0.66 & -- & -- & -- & -- & -- \\
\rowcolor{tabbase} ERA5-Land & 0.56 & 0.051 & 0.62 & -- & -- & -- & -- & -- \\
\bottomrule
\end{tabular}
}
\end{table*}

\begin{table*}[t]
\centering
\caption{\textbf{Observation forecasts and action grounding on the test year.} Columns as in \cref{tab:real_obs}.}
\label{tab:test_obs}
\small
\resizebox{\textwidth}{!}{% generated by code/scripts/make_real_table.py -- do not edit by hand
\begin{tabular}{l cc ccc ccc}
\toprule
\rowcolor{tabhead}  & \multicolumn{2}{c}{Rollout NLL} & \multicolumn{3}{c}{SMAP $\Delta$NLL} & \multicolumn{3}{c}{CYGNSS $\Delta$NLL} \\
\cmidrule(lr){2-3}\cmidrule(lr){4-6}\cmidrule(lr){7-9}
\rowcolor{tabhead} Model & SMAP & CYGNSS & zero & shuffle & swap & zero & shuffle & swap \\
\midrule
\rowcolor{tabours} \method & 5.54$\pm$0.02 & 2.09$\pm$0.00 & 3.39$\pm$0.13 & 1.32$\pm$0.02 & 2.52$\pm$0.04 & 0.36$\pm$0.01 & 0.12$\pm$0.00 & 0.26$\pm$0.00 \\
\quad GRU transition & 5.56$\pm$0.02 & 2.12$\pm$0.01 & 3.03$\pm$0.09 & 1.28$\pm$0.00 & 2.29$\pm$0.04 & 0.29$\pm$0.01 & 0.11$\pm$0.00 & 0.22$\pm$0.00 \\
\quad neural decoder & 5.50$\pm$0.02 & 2.10$\pm$0.00 & 3.60$\pm$0.01 & 1.39$\pm$0.01 & 2.76$\pm$0.05 & 0.34$\pm$0.01 & 0.11$\pm$0.00 & 0.26$\pm$0.01 \\
\quad no action & 6.15$\pm$0.07 & 2.15$\pm$0.01 & -- & -- & -- & -- & -- & -- \\
\quad no rollout loss & 7.34$\pm$0.72 & 2.50$\pm$0.13 & 3.53$\pm$0.17 & 0.99$\pm$0.19 & 1.96$\pm$0.31 & 0.38$\pm$0.19 & 0.09$\pm$0.03 & 0.20$\pm$0.06 \\
\quad black box & 5.48 & 2.11 & 3.25 & 1.37 & 2.45 & 0.30 & 0.11 & 0.23 \\
\rowcolor{tabbase} History Transformer & 6.32 & 2.20 & 3.91 & 1.75 & 3.42 & 0.30 & 0.11 & 0.26 \\
\rowcolor{tabbase} \quad best validation checkpoint & 5.83 & 2.29 & 3.90 & 1.47 & 3.00 & 0.35 & 0.10 & 0.25 \\
\bottomrule
\end{tabular}
}
\end{table*}

\begin{table*}[t]
\centering
\caption{\textbf{Held-out sensor on the test year.} Columns as in \cref{tab:heldout}.}
\label{tab:test_heldout}
\small
% generated by code/scripts/make_heldout_table.py -- do not edit by hand
\begin{tabular}{l c ccc c}
\toprule
\rowcolor{tabhead} Assimilated & seeds & $R$ & ubRMSE & anomaly $R$ & 10-day forecast $R$ \\
\midrule
\rowcolor{tabours} SMAP and CYGNSS & 3 & 0.71$\pm$0.01 & 0.038$\pm$0.002 & 0.76$\pm$0.01 & 0.66$\pm$0.01 \\
SMAP only & 3 & 0.71$\pm$0.01 & 0.038$\pm$0.002 & 0.76$\pm$0.01 & 0.67$\pm$0.01 \\
CYGNSS only & 3 & 0.68$\pm$0.01 & 0.039$\pm$0.002 & 0.72$\pm$0.01 & 0.65$\pm$0.01 \\
\bottomrule
\end{tabular}

\end{table*}

\section{Implementation details}
\label{sec:supp_impl}
\paragraph{Architecture.}
\cref{tab:supp_hparams} summarizes the configuration.
The latent has 5 physical storages (in mm) and 96 free channels.
The transition predicts 12 flux logits per cell; softmax fractions of the available water keep storages non-negative and fluxes storage-limited.
Free channels use leaky memories whose decay time scales are initialized log-uniformly between 6\,h and 365\,d.
A depthwise $5\times5$ convolution followed by a pointwise convolution provides lateral exchange between free channels.
The assimilator embeds tokens into 128 dimensions and runs 4-head cross-attention restricted to each token's cell, followed by two $3\times3$ convolutions that spread the update.
The update is gated; the gate is a sigmoid with a $-2$ offset, so updates start small.
Decoders add a neural residual, scaled by 0.1, to the physical forward models.
All likelihoods are Student-$t$ with $\nu=5$.

\begin{table*}[t]
\centering
\caption{\textbf{Training configuration.}}
\label{tab:supp_hparams}
\small
\begin{tabular}{lll}
\toprule
\rowcolor{tabhead} & Twin experiment & Real data \\
\midrule
Parameters & \multicolumn{2}{l}{\method 1.18\,M (transition 0.54\,M, assimilation 0.53\,M,} \\
 & \multicolumn{2}{l}{decoders 0.05\,M); + latent rain 1.27\,M; History Transformer 4.1\,M} \\
Time step / window & \multicolumn{2}{l}{6\,h; filter 40 steps (10\,d), burn-in 8 steps} \\
Rollout curriculum & \multicolumn{2}{l}{4 $\to$ 40 steps over the first 60\% of training} \\
Optimizer & AdamW, lr $3\cdot10^{-4}$ & AdamW, lr $6\cdot10^{-4}$ \\
 & \multicolumn{2}{l}{weight decay $10^{-4}$, 200-step warm-up, cosine decay, grad.\ clip 1.0} \\
Batch / iterations & 4 windows / 8000 & 32 windows / 6000 \\
Hardware & 1 RTX 4090 & 4 RTX 4090 (data parallel) \\
Sensor dropout & \multicolumn{2}{l}{each sensor with $p=0.2$ per batch} \\
Latent-rain priors & $\sigma_a=0.7$, rain mean 2\,mm, $\sigma_S=0.1$ & not used \\
Memory & \multicolumn{2}{l}{activation checkpointing per time step} \\
\bottomrule
\end{tabular}
\end{table*}

\paragraph{History Transformer baseline.}
Like GraphDOP~\cite{alexe2024graphdop}, this baseline maps past observations to future observations without a persistent physical state.
It forms one token per $8\times8$ patch and time step.
Each token combines the scatter-mean of observation embeddings in the patch with precipitation, context, static fields and positional embeddings.
Six layers of factorized attention (spatial, then causal temporal; width 192, 6 heads) process the tokens.
A query decoder reads the patch token at $t-1$ to predict each observation at $t$, given the same geometry, temperature and soil inputs as our decoders.
During rollout its tokens carry precipitation and context but no observations, and predictions are not fed back.

\paragraph{Compute.}
In the twin, each run uses a single NVIDIA RTX 4090.
\method takes 2--3.5\,s per iteration on the $64\times64$ twin domain; the cost grows with the rollout length, and wall-clock time is dominated by the sequential loop over time steps.
The History Transformer takes 0.3--0.5\,s per iteration.
On real data, each run uses four RTX 4090 with data parallelism and takes 4.5--12.6\,h for the \method variants and 3.0\,h for the History Transformer.

\end{document}